%% file: main.tex
\documentclass[conference]{IEEEtran}
\usepackage{times}

\usepackage{soul,xcolor}
\usepackage[numbers]{natbib}
\usepackage{multicol}
\usepackage{graphicx} 
\usepackage{amsmath,amsthm,amssymb}
\usepackage[noend]{algpseudocode}
\usepackage{amsfonts}
\usepackage{amsmath}
\usepackage{float}
\usepackage{mathrsfs}
\usepackage{booktabs}

\let\labelindent\relax

\usepackage{enumitem}
\usepackage{algorithm}
\usepackage{algpseudocode}
\usepackage{multirow}
\usepackage{stfloats}
\usepackage{etoolbox}
\usepackage{caption}
\usepackage{capt-of}
\usepackage{duckuments}
\usepackage{makecell}

\usepackage[usenames,dvipsnames,table]{xcolor}
\usepackage[bookmarks=true]{hyperref}
\hypersetup{
    colorlinks=true,
    linkcolor=MidnightBlue,
    filecolor=MidnightBlue,      
    urlcolor=MidnightBlue,
    citecolor=MidnightBlue,
}
\usepackage{cleveref}

\setstcolor{blue}

\input{floating/page_wide_teaser}

\begin{document}

\newcommand{\papername}{HydroShear}

\title{HydroShear: Hydroelastic Shear Simulation for Tactile Sim-to-Real Reinforcement Learning}

\author{
    \authorblockN{
        \textbf{An Dang}$^*$$^\dagger$ \quad
        \textbf{Jayjun Lee}$^*$$^\dagger$ \quad
        \textbf{Mustafa Mukadam}$^\ddagger$ \quad
        \textbf{X. Alice Wu}$^\ddagger$ \quad
        \textbf{Bernadette Bucher}$^\dagger$
    }
    \authorblockN{
        \textbf{Manikantan Nambi}$^\ddagger$ \quad
        \textbf{Nima Fazeli}$^\dagger$$^\ddagger$
    }
    \authorblockA{$^\dagger$Robotics Department, University of Michigan $^\ddagger$Amazon IRG (Industrial Robotics Group)}
    \authorblockA{$^*$Equal Contribution}
    \vspace{1mm}
    \authorblockA{
        \textbf{\textcolor{MidnightBlue}{\href{https://hydroshear.github.io/}{hydroshear.github.io}}}
    }
}

\maketitle
\IEEEpeerreviewmaketitle

\input{sections2/0_abstract}

\input{sections2/1_introduction}
\input{sections2/2_related_works}
\input{sections2/3_methodology2}

\input{sections2/4_experiments}
\input{sections2/5_discussion_and_limitations}

\newpage
\clearpage

\section*{Acknowledgments}
\input{sections/acknowledgement}

\bibliographystyle{unsrtnat}
\bibliography{references}

\clearpage
\newpage
\input{sections2/6_appendix}

\end{document}

%% file: floating/page_wide_teaser.tex
\makeatletter
\apptocmd{\@maketitle}{
    \centering
    \includegraphics[width=\textwidth]{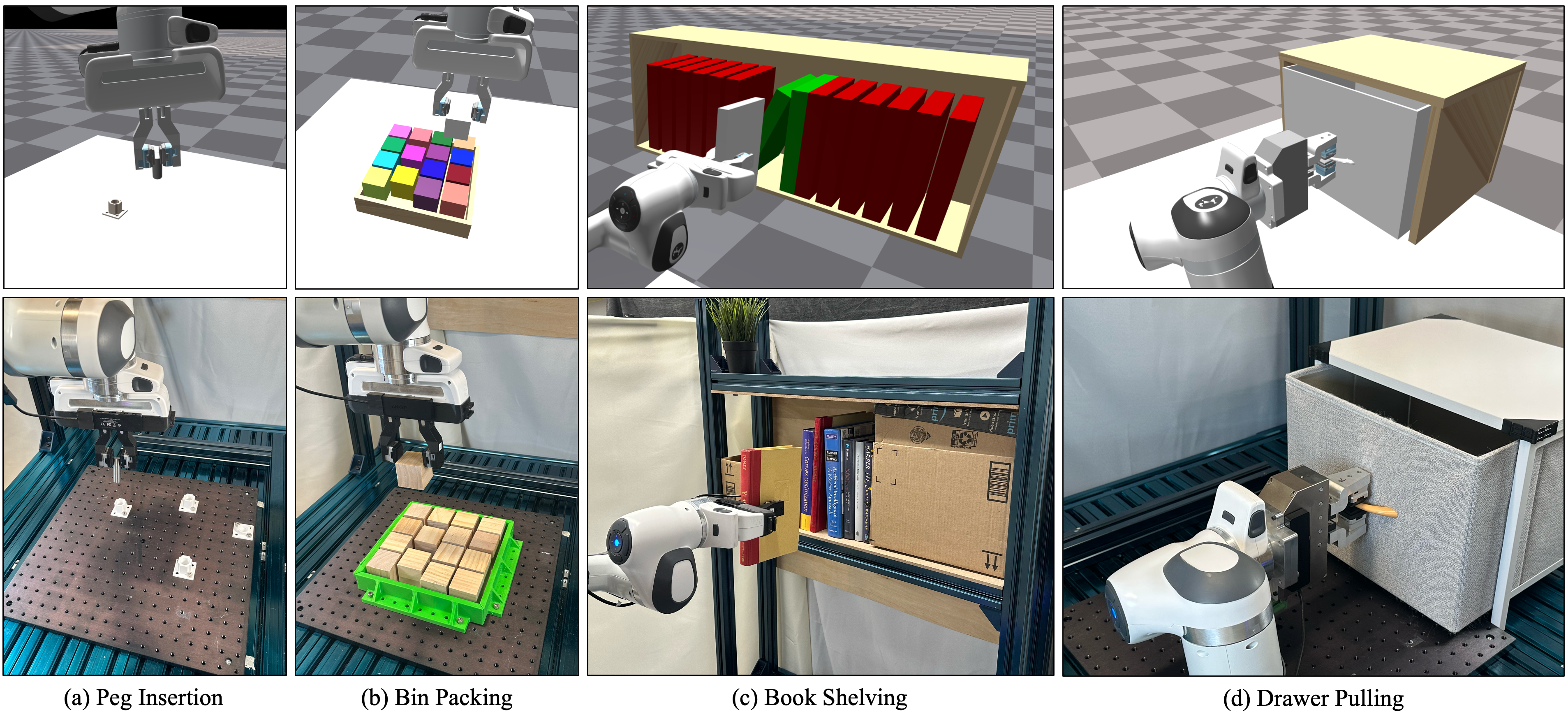}
    \captionof{figure}{\textbf{Illustration of our real-world manipulation tasks.} We present \textbf{HydroShear}, a non-holonomic \underline{\smash{hydro}}elastic model for high-fidelity tactile \underline{\smash{shear}} simulation, for training zero-shot sim-to-real tactile reinforcement learning policies. Our model enables a diverse range of contact and force-rich manipulation tasks that require reasoning over extrinsic contacts and slippages through accurate fingertip tactile shear simulation. Note how each task highlights a different challenge in tactile-based contact-rich manipulation. In (a), the robot must reason about in-hand pose uncertainty of the peg during tight insertion. (b) requires reasoning over simultaneous or sequential multi-object contacts through pushing to insert. (c) demands lateral insertion with gravity acting orthogonal to the insertion axis with full contact patches at the fingertips that make object features such as edges less informative for pose inference. (d) drawer opening tests precise and minimal gripper force modulation given a drawer under unknown force perturbations that induce slippage.}
    \label{fig:page_wide_teaser}
    \setcounter{figure}{1}
}{}{}
\makeatother

%% file: sections2/0_abstract.tex
\begin{abstract}
In this paper, we address the problem of tactile sim-to-real policy transfer for contact-rich tasks. Existing methods primarily focus on vision-based sensors and emphasize image rendering quality while providing overly simplistic models of force and shear. Consequently, these models exhibit a large sim-to-real gap for many dexterous tasks. Here, we present HydroShear, a non-holonomic hydroelastic tactile simulator that advances the state-of-the-art by modeling: a) stick-slip transitions, b) path-dependent force and shear build up, and c) full $\text{SE}(3)$ object-sensor interactions. HydroShear extends hydroelastic contact models using Signed Distance Functions (SDFs) to track the displacements of the on-surface points of an indenter during physical interaction with the sensor membrane. Our approach generates physics-based, computationally efficient force fields from arbitrary watertight geometries while remaining agnostic to the underlying physics engine. In experiments with GelSight Minis, HydroShear more faithfully reproduces real tactile shear compared to existing methods. This fidelity enables zero-shot sim-to-real transfer of reinforcement learning policies across four tasks: peg insertion, bin packing, book shelving for insertion, and drawer pulling for fine gripper control under slip. Our method achieves a 93\% average success rate, outperforming policies trained on tactile images (34\%) and alternative shear simulation methods (58\%–61\%).

\end{abstract}

%% file: sections2/1_introduction.tex
\section{Introduction}

The sim-to-real paradigm shift that has transformed robot locomotion \cite{kumar2021rma, li2021reinforcementbiped, locomamba, diffsimrlquadruped} has yet to materialize for contact-rich manipulation. The primary bottleneck is tactile sensing: a modality that is critical to manipulation but difficult to simulate. In fact, the sim-to-real transfer of reinforcement learning (RL) policies trained with tactile data remains challenging due to discrepancies between simulated contact dynamics and the high-fidelity tactile feedback induced by real contact events. Towards addressing this challenge, a dichotomy has emerged in tactile simulation: visual fidelity has outpaced dynamic fidelity for vision-based tactile sensors.

Recent works have successfully simulated RGB images from vision-based sensors, enabling zero-shot policy deployment for some manipulation tasks \cite{taxim, tacsl, tacex, phong, phong-curved}. However, these images primarily encode contact geometry and fall short of modeling tactile shear~\cite{simshear,lepora_shear}, which captures the force interactions between the sensor and the object. Existing attempts to simulate shear often struggle to bridge the reality gap \cite{tacsl, inhandtranslation, tactilesim, fots}. While Finite Element Methods (FEM) are accurate, they are computationally too expensive for scalable Reinforcement Learning (RL) \cite{narang2020interpreting, ding2020sim}. Conversely, faster penalty-based approaches often approximate shear based merely on instantaneous penetration and velocity, failing to capture the \textit{tactile shadowing} effects~\cite{nisp, oller2023tactilevad}, where deformation induced by one contact region masks or distorts shear information in neighboring regions, or the path-dependent deformation of elastomers of the tactile sensors \cite{tactilesim, tacsl, inhandtranslation}. Even recent marker-based methods such as FOTS are limited to tracking object motion in $\mathrm{SE}(2)$, neglecting full $\mathrm{SE}(3)$ complexity required for dexterous manipulation \cite{fots}.

To address these limitations, we introduce HydroShear, a non-holonomic hydroelastic model for high-fidelity tactile simulation. HydroShear leverages Signed Distance Functions (SDFs) to represent complex geometries and implements a path-dependent contact model that tracks the object's motion history across the sensor membrane during physical interactions. This formulation allows us to simulate realistic shear force fields arising from full $\mathrm{SE}(3)$ motions, effectively capturing complex behaviors like stiction, slippage, and hysteresis. Our approach is  computationally efficient via GPU-parallelization while remaining agnostic to the underlying physics engine. We evaluate our method by training RL policies entirely in simulation and comparing the success rate when zero-shot transferring these policies on dense insertion and fine gripper control tasks. We show that HydroShear's force-aware representation enables a 32\% increase in task success compared to the best performing baseline.

\vspace{-5pt}

%% file: sections2/2_related_works.tex
\section{Related Work}

\noindent\textbf{Tactile Simulation.} Tactile sensing requires simulation of physical contact interactions between the sensor and object. While tactile shear forces can be accurately simulated using FEM~\cite{narang2020interpreting,narang2021sim, ding2020sim}, it is computationally expensive, limiting scalability for training policies at scale. Recent work on FEM-based tactile simulation improves simulation speed via a superposition method to approximate FEM~\cite{taxim}. A new accelerated and parallelized physics simulator, DiffTaichi~\cite{hu2019difftaichi}, also enabled further accelerated FEM-based tactile simulation~\cite{si2024difftactile}. 

Physical approximations of contact interactions have substantially higher computation speed than FEM-based approximations at the cost of fidelity, so this line of tactile simulation research focuses on preserving speed while increasing the accuracy of the simulation. Current simulations perform these physical approximations with rigid body contact models~\cite{tacto}, soft contact, penalty-based models~\cite{tacsl}, and hydroelastic models~\cite{pressurefieldcontact, hydrosoft, bubbles_hydroelastic}. However, these physical approximations struggle to bridge the sim-to-real gap because they do not model tactile shadowing effects such as sensor inaccuracies or signal degradation. Other physical approximations of contact used in sim-to-real transfer~\cite{tactilesim, vitascope} bypass this problem by normalizing the shear field at the cost of losing magnitudinal information. A recent simulator, FOTS~\cite{fots}, uses learning based methods to model tactile shadowing and demonstrates effective sim-to-real transfer. However, FOTS can only simulate $\text{SE}(2)$ motion and cannot simulate shear induced by $\text{SE}(3)$ motions such as rolling.

Our method, \papername{}, models physical interactions with a hydroelastic model. To match our simulation of tactile shear with real-world marker displacement fields, we track the full $\text{SE}(3)$ motion of SDF-based complex geometries relative to the sensor elastomer, enabling zero-shot sim-to-real transfer. Our formulation is GPU-parallelizable, enabling efficient large-scale policy training in simulation. We also provide a GPU-parallelizable implementation of FOTS~\cite{fots}, whose original implementation was not GPU-parallelized, thereby enabling enables efficient large-scale policy training using FOTS.

\noindent\textbf{Tactile Sim-to-Real Transfer and RL.} Sim-to-real transfer of RL policies trained with tactile data remains challenging due to discrepancies between simulated contact dynamics and the high-fidelity tactile feedback induced by real contact events. For tasks involving contact solely between the hand and the object, discretizing or normalizing tactile signals can be sufficient to bridge the sim-to-real gap~\cite{inhandtranslation,tactilesim}. However, for more general contact-rich manipulation tasks such coarse representations no longer suffice. Fine-grained tactile feedback is critical for executing tasks with hand-object-environment interactions with both intrinsic and extrinsic contacts such as peg insertion. Prior work demonstrates successful sim-to-real transfer of RL policies trained on more accurate simulated tactile data for peg insertion~\cite{tacsl,fots}. A differential simulation built on top of TacSL's~\cite{tacsl} contact force estimation model shows results on a broader set of tasks~\cite{si2024difftactile} to ensure generality of the sim-to-real results. We perform sim-to-real transfer of RL policies trained on simulated tactile data from TacSL~\cite{tacsl}, FOTS~\cite{fots}, and \papername{} (ours) on multiple tasks with hand-object-environment interactions (peg insertion, bin packing, book shelving, and drawer pulling). We find that \papername{} substantially outperforms these baselines~\cite{tacsl,fots} in zero-shot sim-to-real transfer.

%% file: sections2/3_methodology2.tex
\section{Methodology}
\input{floating/model_arch_and_distillation_framework}

Our overarching goal is to train tactile-driven manipulation policies in simulation then sim-to-real transfer these policies as in Fig.~\ref{fig:arch_and_disillation}. To achieve this, we present HydroShear, which simulates tactile shear fields induced by the physical interaction between the soft \textit{elastomer} of tactile sensors on the robot and \textit{indenter} objects. This shear field serves as a proxy for the forces transmitted between the robot and the object.

Formally, we define a list of $N$ tactile grid query points (marker positions) on the elastomer membrane $\{ \mathbf{p}_i \}_{i=1}^N$, where $\mathbf{p}_i \in \mathbb{R}^3$, and their associated 2D tactile shear vectors $\{ \mathbf{s}_i \}_{i=1}^N$, where $\mathbf{s}_i = (d_x, d_y) \in \mathbb{R}^2$. While the sensor membrane will undergo a 3D deformation when it is interacted with, our goal is to derive the 2D projection $\mathbf{M}$ of this deformation onto the sensor image plane. $\mathbf{M}$ maps the query points and a sequence of indenter poses in the elastomer frame $\{ {}^E\mathbf{X}^I_t \}_{t=0}^T \in \mathrm{SE}(3)$ to the marker displacement field:
\begin{equation}
\mathbf{M}: \left((x,y), \{ {}^E\mathbf{X}^I_t \}_{t=0}^T \right) \rightarrow \mathbf{s}_i
\end{equation}
where $(x,y) \in \mathbb{R}^2$ represent the coordinates of the tactile query point $\mathbf{p}_i$ on the 2D sensor plane. More concretely, for a vision-based tactile sensor, each image is a collection of pixels that are samples of $(x,y)$ with the corresponding shear $(d_x,d_y)$. $\mathbf{M}$ can be decomposed into dilation and shear fields:
\begin{equation}
    \mathbf{M}_t((x,y), {}^E\mathbf{X}^I_{0:t}) = \mathbf{M}^d_t((x,y)) + \mathbf{M}^s_t((x,y), {}^E\mathbf{X}^I_{0:t})
\end{equation}
where $\mathbf{M}_t$ is the total marker displacement field at time $t$, $\mathbf{M}^d_t$ is the dilation field, and $\mathbf{M}^s_t$ is the shear field. Dilation is due to penetration along sensor normal. Shear is due to object's tangential translation or $\text{SO}(3)$ rotation while in contact and depends on the indenter pose ${}^E\mathbf{X}^I_{0:t}$ history.

\input{floating/projection_figure.tex}

Next, we detail the computation of dilation and shear vector fields (Fig.~\ref{fig:projection}), followed by the system identification procedure used to calibrate the model for a tactile sensor. Finally, we outline the integration of our model into a RL framework to train and sim-to-real transfer tactile policies.

\subsection{Dilation}

We model the dilation vector field by aggregating the influence of all tactile grid query points $\{ \mathbf{p}_i \}_{i=1}^N$ from Fig.~\ref{fig:projection}(b) that are in contact, as shown in Fig.~\ref{fig:projection}(c). We define a tactile grid point $\mathbf{p}_i = \begin{bmatrix} \mathbf{p}_i^{xy} & p_i^z \end{bmatrix}^T$ to be in-contact if $\phi_I(\mathbf{p}_i) < 0$ where $\phi_I: (x,y,z) \in \mathbb{R}^3 \rightarrow \mathbb{R}$ is the signed distance function of the indenter $I$ as shown in Fig.~\ref{fig:projection}(c). Next, we denote the sequence of tactile grid point indices in contact as $C = \{c_i \in \mathbb{N} \ | \ \phi_I(\mathbf{p}_{c_i}) < 0 \}$. With this, we can model the dilation vector field as also seen in FOTS \cite{fots}:
\begin{align}\label{eqn:dilation-vf}
    \mathbf{M}^d_t &= \sum_{i=1}^{|C|} -\phi_I(\mathbf{p}_{c_i}) \cdot \mathbf{v}_{c_i} \cdot \exp\{ -\lambda_d \| \mathbf{v}_{c_i} \|_2^2 \} \\
    \mathbf{v}_{c_i} &= \begin{bmatrix} x \\ y \end{bmatrix} - \mathbf{p}_{c_i}^{xy}
\end{align}
where $\exp\{ -\lambda_d \| \mathbf{v}_{c_i} \|_2^2 \}$ dissipates the effect of the contact point on the queried tactile 2D grid coordinate depending on how far away the point is to model the tactile shadowing effects in the real-world caused by elastomer deformations.

\subsection{Shear}

\input{sections2/3_methodology_forcetrackerv1}

\subsection{Calibration}
\label{method:calib}

We calibrate our model parameters one at a time by isolating the effects of each parameter in the real-world data. Our model contains 4 parameters in total $(\lambda_d, \lambda_s, K,\mu)$. These parameters are calibrated individually in the order they are presented. First we calibrate $\lambda_d$ by collecting tactile shear data $\mathbf{Y}^d_i$ in the real world where the indenter presses down on various positions on the elastomer and formulate the following least-squares optimization problem:
\begin{align*}
    \lambda_d^\star = \text{arg} \min_{\lambda_d} \sum_{i=1}^n \| \mathbf{Y}^d_i - \mathbf{M}^d_t((x_i,y_i); \lambda_d) \|_2^2
\end{align*}

Next we calibrate $\lambda_s$ using tactile shear $\mathbf{Y}^s_i$, but we scale $\mathbf{Y}^s_i$ to be in the same range of vector magnitudes as $\mathbf{M}_t(x,y;\lambda_d^\star,\lambda_s,K,\mu)$. We collect $\mathbf{Y}^s_i$ by dilating the elastomer with the indenter and then translating while the indenter is in contact. This requires us to use the full model and previously calibrated parameters to estimate the vector field as follows:
\begin{align*}
    \lambda_s^\star = \text{arg} \min_{\lambda_s} \sum_{i=1}^n \| \hat{\mathbf{Y}}^s_i -  \mathbf{M}_t((x_i,y_i);\lambda_d^\star,\lambda_s,1,1e5)\|_2^2
\end{align*}
where $\hat{\mathbf{Y}}^s_i$ is the rescaled real shear vector field. Afterwards, we calibrate $K$ but without the rescaling trick:
\begin{align*}
    K^\star = \text{arg} \min_{K} \sum_{i=1}^n \| \mathbf{Y}^s_i -  \mathbf{M}_t((x_i,y_i);\lambda_d^\star,\lambda_s^\star,K,1e5)\|_2^2
\end{align*}

Finally, we calibrate $\mu$ by collecting data of when the indenter is slipping on the elastomer and solve the following problem:
\begin{align*}
    \mu^\star = \text{arg} \min_{\mu} \sum_{i=1}^n \| \mathbf{Y}^\mu_i -  \mathbf{M}_t((x_i,y_i);\lambda_d^\star,\lambda_s^\star,K^\star,\mu)\|_2^2
\end{align*}

In summary, we calibrate the dilation dissipation parameter $\lambda_d$, tangential shear dissipation parameter $\lambda_s$, tangential stiffness parameter $K$, and friction coefficient $\mu$. We solve individual single-variable optimization problems to improve the conditioning of the calibration procedure and avoid directly solving the highly nonlinear joint optimization, for which reliable convergence to good optima is difficult to guarantee.

\subsection{Sim-to-Real Reinforcement Learning}
\label{sim2real-method}
We formulate our contact-rich manipulation tasks with tactile feedback as infinite-horizon discrete-time Partially-Observable Markov Decision Processes defined by the tuple $\mathcal{M}:=( \mathcal{S}, \mathcal{O}, \mathcal{A}, \mathcal{R}, \mathcal{P}, \rho_0, \gamma )$ with state space $\mathcal{S}$, the observation space $\mathcal{O}$, action space $\mathcal{A}$, reward $\mathcal{R}(s,a)$, transition probability $\mathcal{P}(s'|s,a)$, initial state distribution $\rho_0$, and discounting factor $\gamma \in [0,1)$. The robot seeks to learn a parameterized policy $\pi_\theta(\cdot|s)$ to maximize the expected discounted return $\mathcal{J}:=\mathbb{E}_{\pi_\theta}\left[ \sum_{k=0}^{\infty} \gamma^k r_{t+k} \mid s_t = s \right]$. The only distinction between simulation and real is access to the state space. \\
\noindent\textbf{Teacher-Student Distillation:} We train HydroShear RL policies entirely in Isaac Gym~\cite{isaacgym} simulation with parallelized environments using PPO~\cite{schulman2017proximal} and zero-shot deploy in the real-world without any modification or fine-tuning. We list the task reward functions in Appx.~\ref{app:rewards}. As outlined in Fig.~\ref{fig:arch_and_disillation}, we first train a teacher actor-critic using privileged state information. We find that naively training RL agents with contact penalties results in suboptimal policy that avoids contact with the goal, thus preventing successful task completion. To prevent such collapse, we design a teacher training curriculum where we first start training with no contact penalty until initial task success. Then, we finetune with the contact penalty term in the next stage. Next, we follow asymmetric actor-critic distillation (AACD)~\cite{tacsl} where we use the privileged teacher critic to train a student policy actor from scratch. Note that all students for each task are trained with the same teacher critic checkpoint trained on the contact penalty curriculum. For a more detailed descriptions of the training pipeline, refer to Appx.~\ref{appendix:training_details}. Our student policies observe their proprioceptive states, goal pose, and tactile feedback with no access to privileged information such as in-hand object poses or contact forces.

%% file: floating/model_arch_and_distillation_framework.tex
\begin{figure*}[!ht]
    \centering
    \includegraphics[width=0.95\linewidth]{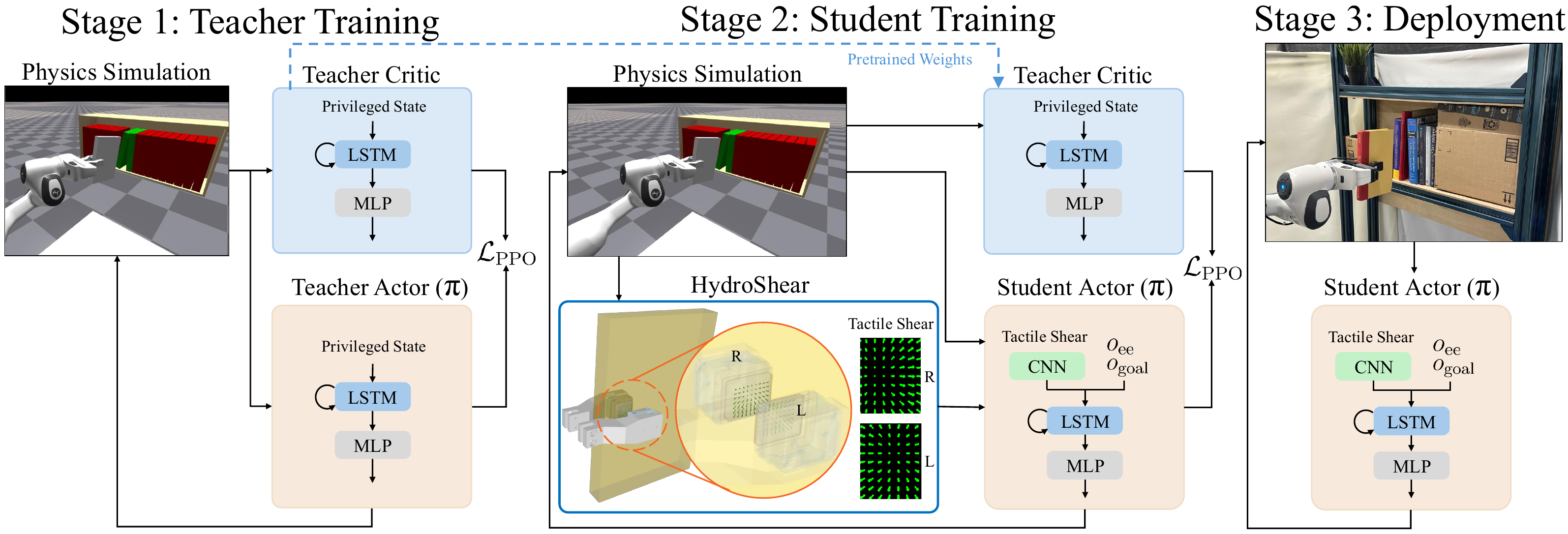}
    \caption{\textbf{Illustration of the teacher-student RL training using Asymmetric Actor-Critic Distillation (AACD)}~\cite{tacsl}. Stage 1: a teacher actor-critic is trained with access to privileged states such as contact forces and object poses. Stage 2: the critic is initialized with the expert critic pretrained from Stage 1 and the student actor that takes high-dimensional inputs (end-effector (EE) pose, relative EE-goal pose, left and right tactile shear) is trained from scratch. The actor-critic is optimized with the PPO RL objective and uses encoder-LSTM-MLP networks (see Appx.~\ref{appendix:network_arch}). Stage 3: we deploy the student actor in the real world.}
    \label{fig:arch_and_disillation}
    \vspace{-6mm}
\end{figure*}

%% file: floating/projection_figure.tex
\begin{figure*}[!ht]
    \centering
    \includegraphics[width=0.9\linewidth]{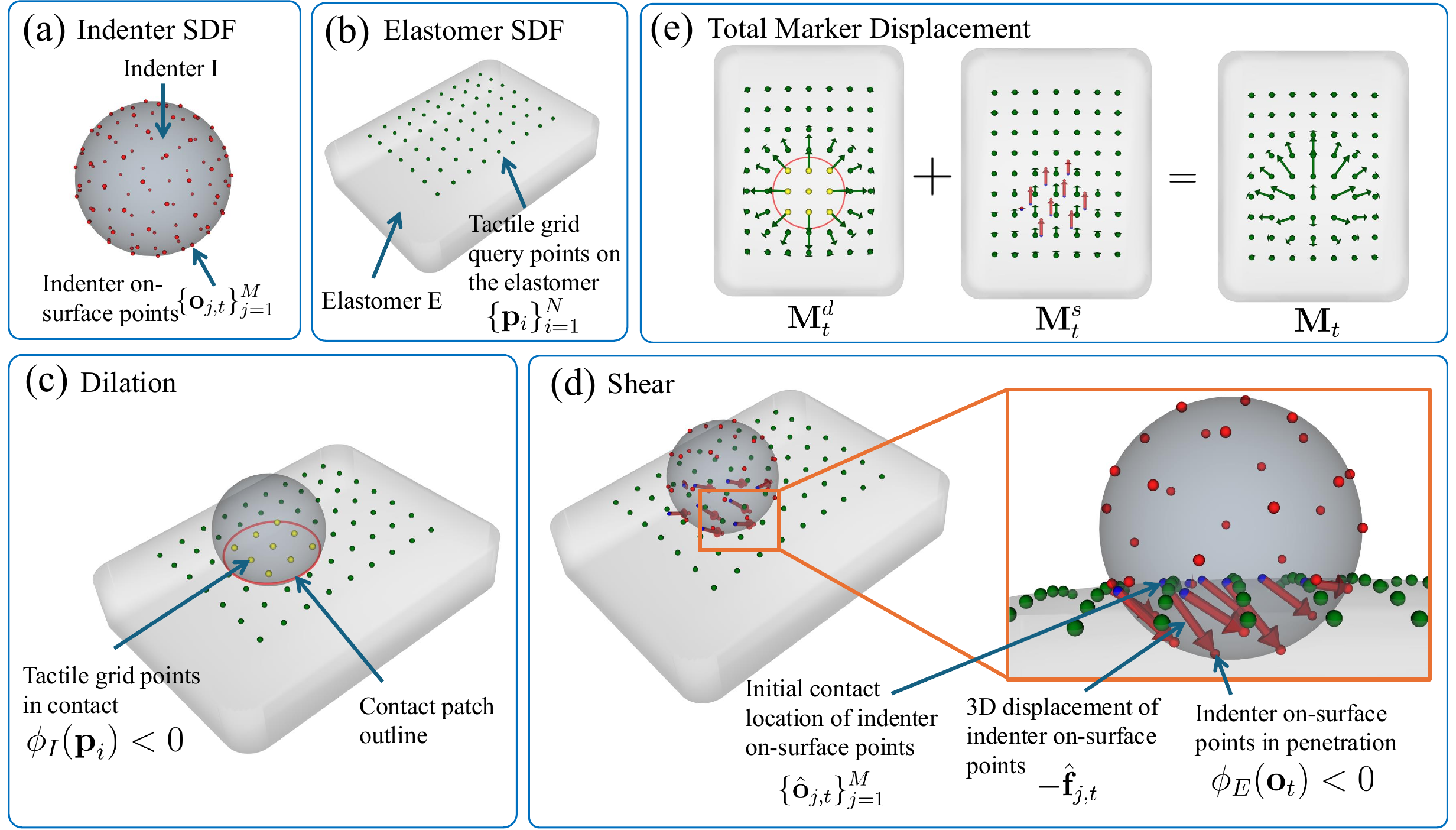}
    \caption{\textbf{Illustration of \papername{} and its pipeline for tactile shear simulation.} \papername{} simulates the tactile shear feedback that arises from the physical interaction between the indenter $I$ in (a) and the sensor elastomer $E$ in (b). The goal is to compute the marker displacement fields across the tactile grid query points on the elastomer $\{ \mathbf{p}_i \}_{i=1}^N$ from (b). \papername{} computes the dilation displacement field $\mathbf{M}_t^d$ in (c) and the shear displacement field $\mathbf{M}_t^s$ in (d) to get the total marker displacement field $\mathbf{M}_t$ in (e). The dilation field is computed by identifying the tactile grid points that are in contact with the indenter SDF ($\phi_I(\mathbf{p}_i) < 0$). Here, the red circle represents the outline of the contact patch. For the shear field, we take the history of indenter poses in the elastomer frame ${}^E\mathbf{X}^I_{0:t}$ to track the 3D displacement $-\hat{\mathbf{f}}_{j,t}$ of the indenter on-surface points $\{\mathbf{o}_{j,t}\}_{j=1}^M$, represented by the red arrows in (d), which connects the initial contact location of indenter on-surface points ($\{\mathbf{\hat{o}}_{j,t}\}_{j=1}^M$) to the current position of the on-surface indenter points while in-penetration to the elastomer SDF ($\phi_E(\mathbf{o}_t) < 0$).
    }
    
    \label{fig:projection}
    \vspace{-15pt}
\end{figure*}

%% file: sections2/3_methodology_forcetrackerv1.tex
We introduce a hydroelastic model that extends the hydrostatic pressure field contact model~\cite{pressurefieldcontact, bubbles_hydroelastic} to track forceful interactions between the indenter and the sensor's compliant elastomer. Specifically, we track contact forces on a set of indenter on-surface points $\{ \mathbf{o}_{j,t} \}_{j=1}^M$ in the elastomer frame where $M$ is the total number of points in Fig.~\ref{fig:projection}(a). The coordinates of $\mathbf{o}_{j,t}$ are calculated by transforming the local indenter surface points in indenter frame $\bar{\mathbf{o}}_{j}$ with the indenter pose ${}^E\mathbf{X}^I_t$ via $\mathbf{o}_{j,t} = {}^E\mathbf{X}^I_t \bar{\mathbf{o}}_{j}$. To account for the viscoelasticity of the elastomer membrane, we recursively track the contact forces $\{ \tilde{\mathbf{f}}_{j,t} \}_{j=1}^M$ by taking the indenter object pose path in elastomer frame $({}^E\mathbf{X}^I_t,{}^E\mathbf{X}^I_{t-1})$ and converting the displacement of the in-contact indenter surface points into forces. We compute this using the viscoelastic stiffness parameters of the elastomer $(E,K, A_j)$ where $E$ is the stiffness along the indenter's normal direction, $K$ is the stiffness tangential to the sensor normal, and $A_j$ is the surface area associated to the indenter surface points. We model our recursive formulation as:
\begin{equation}\label{eqn:track-displacement-hydrosoft}
    \tilde{\mathbf{f}}_{j,t} = F(\tilde{\mathbf{f}}_{j,t-1}, {}^E\mathbf{X}^I_t, {}^E\mathbf{X}^I_{t-1}, \bar{\mathbf{o}}_j; E, K, A_j, \mu)
\end{equation}
where $\mu$ denotes the friction coefficient for the forceful interaction between the indenter and elastomer. For our recursive formulation, we assume in the base case $t=0$ that the indenter and elastomer are not in contact, thus $\tilde{\mathbf{f}}_{j,0} = 0$ for all $j$.

To calculate $\tilde{\mathbf{f}}_{j,t}$ through the force tracking function $F$, our hydroelastic force tracker first takes the displacement of the indenter surface points at time $t$ and $t-1$ and only tracks the portion of this displacement $\mathbf{o}_{j,t} - \mathbf{o}_{j,t-1}$ that is in contact with the elastomer. We approximately determine this portion using the elastomer signed distance function $\phi_E: (x,y,z) \in \mathbb{R}^3 \to \mathbb{R}$ by calculating a fraction $\alpha_{j,t}$ of the SDF difference $\phi_E(\mathbf{o}_{j,t}) - \phi_E(\mathbf{o}_{j,t-1})$ that is in-contact:
\begin{align}\label{eqn:fraction-displacement}
\alpha_{j,t} &= \frac{\text{ReLU}(-\phi_E(\mathbf{o}_{j,t})) - \text{ReLU}(-\phi_E(\mathbf{o}_{j,t-1}))}{\phi_E(\mathbf{o}_{j,t}) - \phi_E(\mathbf{o}_{j,t-1})} \\
\mathbf{d}_{j,t} &= \alpha^d_{j,t}(\mathbf{o}_{j,t} - \mathbf{o}_{j,t-1})
\end{align}
where $\mathbf{d}_{j,t}$ is the indenter surface point displacement scaled by how much of the displacement is in-contact. We introduce the \text{ReLU} into Eq.~\ref{eqn:fraction-displacement} to compactly represent 3 cases that can occur when tracking indenter surface point displacements. When $\phi_E(\mathbf{o}_{j,t})$ and $\phi_E(\mathbf{o}_{j,t-1})$ are both negative, the indenter surface point displacement is fully in contact with the elastomer and $\alpha_{j,t}=1$. When both are positive, indenter surface point displacements are fully out of contact with the elastomer and $\alpha_{j,t} = 0$.
When they have opposite signs, we will get an $\alpha_{j,t}$ such that $0 < \alpha_{j,t} < 1$ because only a part of the indenter surface point displacement will be inside the elastomer.

Using the viscoelastic properties $(E,K)$ of the elastomer, we convert the tracked in-contact indenter surface point displacements $\mathbf{d}_{j,t}$ and convert them into forces $\mathbf{f}_{j,t}$. We decompose $\mathbf{d}_{j,t}$ into a normal component $d^n_{j,t}$ and tangential component $\mathbf{d}^{xy}_{j,t}$. We compute $d^n_{j,t}$ by projecting $\mathbf{d}_{j,t}$ onto the indenter object normal $\hat{\mathbf{n}}_j$ at the indenter surface point $\mathbf{o}_{j,t}$ with $d^n_{j,t} = \langle \mathbf{d}_{j,t}, \hat{\mathbf{n}}_j \rangle$. By subtracting the normal component $d^n_{j,t}\hat{\mathbf{n}}_j$ from the original in-contact displacement $\mathbf{d}_{j,t}$, we recover the tangential component $\mathbf{d}^{xy}_{j,t} = \mathbf{d}_{j,t} - d^n_{j,t} \hat{\mathbf{n}}_j$. With this in mind, we linearly scale the displacements into forces:
\begin{align}\label{eqn:hydrosoft1}
    f^n_{j,t} = \tilde{f}^n_{j,t-1} + EA_j d^n_{j,t} \\
    \mathbf{f}^{xy}_{j,t} = \tilde{\mathbf{f}}^{xy}_{j,t-1} + KA_j \mathbf{d}^{xy}_{j,t}
\end{align}

Next, we impose that normal forces be lower bounded by zero (no adhesion) and enforce Coulomb friction by clipping the magnitude of the tracked tangential friction force $\mathbf{f}_{j,t}^{xy}$ by the normal force scaled by the coefficient of friction:
\begin{align}\label{eqn:hydrosoft2}
    \bar{f}^n_{j,t} = \text{ReLU}(f^n_{j,t}) \\
    \bar{\mathbf{f}}^{xy}_{j,t} = \min \{1, \frac{\mu \bar{f}^n_{j,t}}{\| \mathbf{f}_{j,t}^{xy} \|_2} \} \mathbf{f}_{j,t}^{xy}
\end{align}
where $\mu$ is the friction coefficient between the elastomer and indenter. $\mu$ determines how much tangential force $\mathbf{f}^{xy}_{j,t}$ is required before sliding occurs. Finally, the contact forces are reset if the indenter surface point breaks contact with the elastomer:
\begin{align}\label{eqn:hydrosoft3}
    \tilde{\mathbf{f}}_{j,t} = H(-\phi_E(\mathbf{o}_{j,t})) \cdot (\bar{f}^n_{j,t} \hat{\mathbf{n}}_j + \bar{\mathbf{f}}^{xy}_{j,t})
\end{align}
where $H$ is the Heaviside function.

As part of the shear vector field computation, we take the contact forces $\tilde{\mathbf{f}}_{j,t}$ on the indenter surface and project it onto the elastomer surface.  Specifically, for every indenter surface point $\mathbf{o}_{j,t}$ that is in contact with the elastomer, we want to find the corresponding point on the elastomer surface $\hat{\mathbf{o}}_{j,t}$ that the indenter point is in contact with. This way, each contact force $\tilde{\mathbf{f}}_{j,t}$ has a corresponding point on the elastomer surface that it is being applied on. When the interaction between the indenter and elastomer is sticking, $\hat{\mathbf{o}}_{j,t}$ does not change. However, when the interaction is slipping, $\hat{\mathbf{o}}_{j,t}$ shifts to represent the change in the point of contact on the elastomer surface.

In order to model these effects, we formulate projection as finding a set of displacements $\{ \hat{\mathbf{f}}_{j,t} \}_{j=1}^M$ that are applied onto the indenter surface points $\{ \mathbf{o}_{j} \}_{j=1}^M$ to acquire $\{ \hat{\mathbf{o}}_{j,t} \}_{j=1}^M$:
\begin{align}
    \hat{\mathbf{o}}_{j,t} = \mathbf{o}_{j,t} + \hat{\mathbf{f}}_{j,t}
\end{align}
We illustrate the displacements $-\hat{\mathbf{f}}_{j,t}$ by the red arrows on $\mathbf{M}_t^s$ in Fig.~\ref{fig:projection}(e). To compute $\hat{\mathbf{f}}_{j,t}$ such that it provides slippage and provides tracked indenter surface displacement while in contact, we formulate an indenter displacement tracker based on the Eqn.~\ref{eqn:track-displacement-hydrosoft} but the elastomer stiffness parameters and the surface area parameter $A_j$ for this tracker are set to $1$:
\begin{align}
    \hat{\mathbf{f}}_{j,t} = F(\hat{\mathbf{f}}_{j,t-1}, {}^E\mathbf{X}^I_t, {}^E\mathbf{X}^I_{t-1}, \bar{\mathbf{o}}_j; 1, 1, 1, \hat{\mu})
\end{align}
where $\hat{\mu}$ provides the slippage effect for the projected indenter points on the elastomer surface $\hat{\mathbf{o}}_{j,t}$. 

To calculate the shear vector field, we would need both the hydroelastic contact force tracker and the projection displacement tracker. However, we show that only one tracker is needed and the other can be recovered given a few assumptions. If we set $E=K$, $\mu=\hat{\mu}$, and $A_j=A$ for the hydroelastic contact force tracker where $A$ is a constant that assumes the surface area associated with each indenter surface point is uniform, then we can recover the contact forces from just the displacements by scaling the $\hat{\mathbf{f}}_{j,t}$ by KA as follows:
\begin{align}
    \tilde{\mathbf{f}}_{j,t} = KA \cdot \hat{\mathbf{f}}_{j,t}
\end{align}

After computing $\hat{\mathbf{f}}_{j,t}$ and $\tilde{\mathbf{f}}_{j,t}$, we calculate the shear vector field using the aggregated effects of each contact force $\tilde{\mathbf{f}}_{j,t}$ at the projected indenter surface point $\hat{\mathbf{o}}_{j,t} = \begin{bmatrix} \hat{\mathbf{o}}_{j,t}^{xy} & \hat{o}_{j,t}^z \end{bmatrix}^T$ on each of tactile 2D grid query coordinate. We denote the set of indenter on-surface points that are in contact with the elastomer as $K_t = \{ k_i \space \ | \ \space \phi_E(\mathbf{o}_{k_i,t}) < 0 \}$ in Fig.~\ref{fig:projection}(d). Now, we calculate the shear vector field as follows: 
\begin{align}
    \mathbf{M}_t^s &= \sum_{i=1}^{|K_t|} -\phi_E(\mathbf{o}_{k_i,t}) \cdot -\tilde{\mathbf{f}}^{xy}_{k_i,t} \cdot
    \exp\{-\lambda_s \| \mathbf{v}_{k_i} \|_2^2 \} \\
    \mathbf{v}_{k_i} &= \begin{bmatrix} x \\ y \end{bmatrix} - \hat{\mathbf{o}}^{xy}_{k_i,t}
\end{align}
where we weigh the effect each indenter surface point $\mathbf{o}_{j,t}$ has on the vector field by its penetration depth $-\phi_E(\mathbf{o}_{k_i,t})$ and dissipate the effects of the surface point on a queried tactile 2D grid coordinate $(x,y)$ based on distance using $ \exp\{-\lambda_s \| \mathbf{v}_{k_i} \|_2^2 \}$. In total, our method takes in 4 parameters $(\lambda_d, \lambda_s, K, \mu)$. Next, we present how to perform system identification to simulate realistic tactile shear fields.

%% file: sections2/4_experiments.tex
\section{Experiments and Results}

We evaluate HydroShear along two axes: (1) its ability to model diverse modes of real-world shear induced by full $\text{SE}(3)$ indenter motion with respect to the elastomer, and (2) its ability to zero-shot sim-to-real transfer RL policies. To this end, we first describe the real-world calibration setup used to collect a dataset of tactile shear field for system identification. We then assess HydroShear’s shear simulation accuracy using a separate validation dataset with unseen contact configurations in a digital twin environment, comparing against TacSL~\cite{tacsl} and FOTS~\cite{fots}. Next, we test our sim-to-real policies across four contact rich tasks, consisting of three dense insertion tasks: (1) Peg Insertion, (2) Bin Packing, (3) Book Shelving, and (4) Drawer Pulling that specifically requires precise gripper control and attention to slippage. Additionally, we conduct a speed comparison experiment between FOTS and \papername{} in Tab.~\ref{tab:speed-table}

\subsection{Calibration and Real-world Shear Evaluation}
\input{floating/calibration_setup.tex}

We calibrate our method's parameters (Sec.~\ref{method:calib}) by collecting controlled marker displacement data designed to isolate the contribution of each parameter $(\lambda_d, \lambda_s, K, \mu)$. Fig.~\ref{fig:calibration} shows our experimental setup, were we use a 7-DoF KUKA MED LBR robot equipped with a 35-mm spherical indenter mounted at the tool center point (TCP) to calibrate the sensor. Here, we use the GelSight Mini sensor that we rigidly mount to the table. During calibration, we assume access to the full poses of both the tactile sensor and the sphere indenter. 

To calibrate $\lambda_d$, we collect 10 samples in which the sphere indenter presses vertically into the elastomer at randomly sampled surface locations, and select the value of $\lambda_d$ that best matches the resulting tactile shear response. To calibrate $\lambda_s$ and $K$, we collect 10 additional samples in which the indenter applies combined normal and tangential motions, shearing the elastomer at random locations and directions, and jointly optimize $\lambda_s$ and $K$ to best fit the observed tactile shear. Each sample contains the actual tactile shear field measured as a marker displacement field and the indenter trajectory that induced the shear field.  Finally, to estimate the friction coefficient $\mu$, we collect 10 samples of controlled slip motions while maintaining contact with the elastomer and select $\mu$ to best match the resulting output. To convert the predicted shear displacement from meters to pixels, we use the GelSight Mini camera resolution, yielding a scale factor of $1000 / 0.065 \ \text{[px/m]}$.

\noindent\textbf{Baselines:} We evaluate the performance of our proposed model by comparing it against prior methods that simulate tactile shear and support sim-to-real policy transfer. Specifically, we benchmark against two representative baselines: TacSL~\cite{tacsl} and FOTS~\cite{fots}. TacSL implements a penalty-based force field~\cite{tactilesim} in Isaac Gym, leveraging the object signed distance field (SDF) and the object’s relative velocity with respect to the elastomer to estimate contact forces at tactile grid query points on the elastomer $\{\mathbf{p}_i\}_{i=1}^N$ from Fig.~\ref{fig:projection}(b). FOTS~\cite{fots} decomposes the marker displacement field into three components that track object motion in $\mathrm{SE}(2)$ (see Appx.~\ref{appendix:fots}). For a fair comparison, we follow each method’s original calibration procedure and use the same real-world calibration dataset to estimate their model parameters.

\noindent\textbf{Results:} We evaluate our method against the baselines by comparing the simulated tactile shear field from the digital twin with real-world measurements of marker displacements with Gelsight Minis. We collect 10 samples each for dilation, translational shear, torsional shear (twist), and rolling shear, resulting in a total of 40 test samples. Tab.~\ref{tab:calibration_results} reports the per-taxel Root Mean Squared Error (RMSE) and mean cosine similarity (CS) between the predicted and ground-truth shear vector fields, averaged over all taxels, with qualitative samples shown in Fig.~\ref{fig:vedo_calibration}. Our method achieves consistently lower error and higher directional alignment across all shear types, indicating a more faithful representation of elastomer shear behavior. TacSL exhibits increased error due to the absence of tactile shadowing in its formulation, which limits its ability to accurately capture shear patterns. FOTS explicitly accounts for tactile shadowing and attains comparable performance for in-plane object motions. However, its accuracy degrades under tilt motions, which involve out-of-plane components of $\mathrm{SE}(3)$ that are not explicitly modeled in the FOTS formulation. More broadly, FOTS represents contact through relative object motion expressed in an object-centric frame rather than an SDF-based geometric representation. In contrast, our method leverages object SDFs and on-surface contact points to resolve higher-resolution contact motion and interaction paths, enabling more precise tactile shear simulation, particularly for complex and out-of-plane interactions. Finally, Fig.~\ref{fig:complex_geom} illustrates \papername{}'s ability to simulate complex object geometries involving multiple simultaneous contact patches.

\input{tables/marker_table}

\input{floating/vedo_calibration}
\input{floating/vedo_complex_geometries}

\subsection{Zero-shot Sim-to-Real RL}\label{sec:4B-zero-shot-sim-to-real}

\noindent\textbf{Real-world Setup:} We conduct our sim-to-real experiments using a 7-DoF Franka Emika Panda robot equipped with a pair of GelSight Mini vision-based tactile sensors mounted. We mount the sensors on the default Franka hand gripper for Peg Insertion, Bin Packing, and Book Shelving tasks and on a Weiss WSG-50 gripper for Drawer Pulling task for more precise gripper control (Fig.~\ref{fig:page_wide_teaser}). We compute the real-world shear from GelSight Minis using marker displacements.

The four real-world tasks in Fig.~\ref{fig:page_wide_teaser} are designed to highlight distinct use cases of high-fidelity tactile shear simulation for sim-to-real policy learning without any visual feedback. %

\noindent\textbf{1. Peg Insertion:} This task is designed to introduce in-hand object pose uncertainty during force-rich insertion. The task goal is to insert a cylindrical peg, grasped at varying in-hand poses, into a socket mounted on the tabletop as shown in Fig.~\ref{fig:page_wide_teaser}(a). 
For the real-world rollouts and testing, we tested each policy over 3 different goals with 10 different in-hand orientations in the range of $[-30, 30]$ degrees per goal.

\noindent\textbf{2. Bin Packing:} This task is designed to handle multi-object interactions during an insertion. The goal is to insert a grasped wooden cube block into a specified target location within a 4 by 4 bin populated with other wooden cubes. The cubes inside the bin are arranged in a grid of rows and columns, with orientations aligned to the bin and with uniform spacing between neighboring cubes, as shown in Fig.~\ref{fig:page_wide_teaser}(b). To increase the task difficulty, cubes adjacent to the goal are squished either vertically or horizontally, partially or heavily occluding the goal (see Appx.~\ref{appendix:randomization_modes}), using either one or two cubes.

\noindent\textbf{3. Book Shelving:} This task tests lateral insertion where gravity is no longer exerting forces along the axis of insertion. Note the grasped book is larger than the fingertip and therefore the robot cannot infer the object pose solely from touch. The goal is to insert the book, grasped at a consistent in-hand pose, into a target location on a bookshelf where neighboring books occlude the goal via parallel translation or tilting (see Appx.~\ref{appendix:randomization_modes}), as in Fig.~\ref{fig:page_wide_teaser}(c). We further evaluate performance across books of varying sizes to test more diverse settings.

\noindent\textbf{4. Drawer Pulling:} This task is designed to test gripper width modulation and reaction to slippage by learning the continuous control of the gripper position. The robot starts by grasping the drawer handle at an initial gripper width that is sufficient to pull the drawer but allows slippage under external disturbances or additional load on the drawer. The task goal is to modulate the gripper width only if an external force is sensed or slippage is detected, thereby preventing further slippage while opening the drawer. 
To challenge the grasp stability, we apply multiple force perturbations to the drawer at varying timings during pulling. In our evaluation, we introduce three different weight loads and vary the perturbation position and duration. The drawer handle leaving the grasp is considered a failure.

\noindent\textbf{Baselines:} We compare the zero-shot sim-to-real performance of tactile RL policies each trained with different tactile simulation framework against our method \papername{}. As outlined in Sec.~\ref{sim2real-method}, we first trained a teacher actor-critic for each task through the contact penalty curriculum as shown in Fig.~\ref{fig:rl_training_curves_teacher}. Next, we train one student policy per task and per tactile simulation framework using AACD with using the same teacher checkpoint for each task, as illustrated in Fig.~\ref{fig:arch_and_disillation}. The network architectures (see Appx.~\ref{appendix:network_arch}) are identical across tasks and policies, taking the same set of observations (EE pose, relative EE-goal pose, and tactile feedback), with the only difference being the tactile simulation framework.

Here, we highlight the key distinctions between the different baseline student actors with their tactile feedback. (1) TacSL Gray: We train a tactile image-based policy and use grayscale tactile image as we found it to work well empirically. %
(2) TacSL Shear: We train a TacSL shear-based policy where we normalize the tactile shear into a unit vector, per-taxel, to mitigate for the lack of real-world tactile shadow effects, a strategy adopted by prior works~\cite{tactilesim, tacsl, vitascope}. 
Next, we train two shear-based FOTS policies: (3) Original FOTS: following the original FOTS formulation but \textit{reimplemented by us for parallelization}, and (4) a second \textit{improved, parallelized} variant that additionally computes marker displacements relative to the initial contact patch center between the object and elastomer, rather than the object center under $\mathrm{SE}(2)$ motion that the original FOTS proposed (see Appx.~\ref{appendix:fots}). We show that these modifications improve performance significantly and provide a much stronger baseline to compare against.
Finally, we train (5) our \papername{}-based policies.
The RL training curves for these student policies for each task can be found in Fig.~\ref{fig:rl_training_curves_peg_insertion}-\ref{fig:rl_training_curves_drawer_pulling}.

\noindent\textbf{Gravity Effect on Shear:} For displacement-based shear algorithms like FOTS and HydroShear, we also add in a gravity effect in simulation to compensate for the object's gravity pulling down on the grasped object and hence the elastomer while in-grasp. We discuss this in detail in Appx.~\ref{app:gravity-effect}.

\noindent\textbf{Results: }The experimental results are summarized in Tab.~\ref{tab:sim_to_real_results}. We report success rates over episode rollouts and the time-to-success for successful episodes in Tab.~\ref{tab:real_world_timesteps} in Appx. Fig.~\ref{fig:rollout_keyframes} shows representative sim-to-real rollout keyframes across all tasks along with tactile shear feedback. Across all four real-world tasks, we observe a clear relationship between the fidelity of the simulated tactile shear representation and zero-shot sim-to-real performance in Tab.~\ref{tab:sim_to_real_results}.

\input{tables/zeroshot_table}

TacSL Gray performs moderately on Peg Insertion and Bin Packing, contrasted with substantially poorer transfer on Book Shelving and Drawer Pulling. TacSL Gray is an image-based tactile policy, and such policies perform reasonably well when contact patches are small and localized. For Book Shelving, however, full fingertip contact patches hinder textures and the lack of explicit shear direction, accumulation, and slip cues in Drawer Pulling cause the performance to degrade sharply.

TacSL Shear achieves strong performance across most tasks but degrades notably on Bin Packing. This limitation arises from its reliance on per-taxel shear normalization, which discards critical information about shear magnitude and the spatial structure of contact geometries with tactile shadows. While this normalization is effective for Peg Insertion, where it preserves contact geometry and in-hand pose, and for Drawer Pulling, where the formulation naturally captures slippage, it becomes detrimental for Bin Packing. In this task, large shear and torsional interactions induce broader contact patches and global elastomer deformations producing normalized shear fields that span a much wider patch than what the student policy was trained to expect, leading to reduced performance.

The original FOTS implementation performs poorly on the Peg Insertion and Bin Packing tasks because the object frame center is defined much further away from the contact patch center between the elastomer and grasped objects. Conversely, it performs well on Book Shelving and Drawer Pulling because the object frame center is much closer to the contact patch center between the elastomer and grasped object. FOTS exposes a sim-to-real gap when the object frame is poorly aligned with the grasp or when contact occurs far from the reference origin. In these cases, the resulting tactile signals fail to reflect the local contact dynamics experienced by the sensor. 

FOTS (Reimpl.) substantially improves transfer for tasks with relatively stable contacts, but the performance degrades for Drawer Pulling. Re-centering shear around an initial contact patch enables this improved transfer for stable contacts, but when a single reference patch is no longer well-defined in Drawer Pulling, the performance degrades. The assumption of the contact patch location breaks down under frequent contact transitions, intermittent slip, or external perturbations. %

In contrast, HydroShear achieves robust zero-shot transfer across each tasks by explicitly modeling path-dependent, SE(3) contact interactions by tracking on-surface indenter points. By tracking how shear accumulates, dissipates, and transitions through stick–slip events, the simulated tactile feedback remains consistent with real sensor observations even as contact configurations evolve. This enables a single policy architecture and training pipeline to generalize across dense insertion, multi-object interactions, full-contact manipulation, and slip-sensitive force modulation.

We provide a qualitative comparison of different tactile shear feedbacks during policy rollouts in sim for each task in Fig.~\ref{fig:peginsertion_rollout_keyframes}-\ref{fig:drawerpulling_rollout_keyframes} to compare against real rollouts in Fig.~\ref{fig:rollout_keyframes}.

Ablation on Contact Penalty: We empirically find that teacher checkpoints finetuned with the contact penalty curriculum significantly improve downstream student policy behavior during sim-to-real transfer. For example, peg insertion policies trained without any contact penalty make overly aggressive contact with the socket and repeatedly react in the same manner, damaging the soft elastomer membrane of the GelSight Mini sensors. In contrast, using the contact penalty curriculum yields safer tactile policies that make controlled contact with the environment and exhibit emergent goal-searching behavior.
\par

%% file: floating/calibration_setup.tex
\begin{figure}
    \centering
    \includegraphics[width=0.9\linewidth]{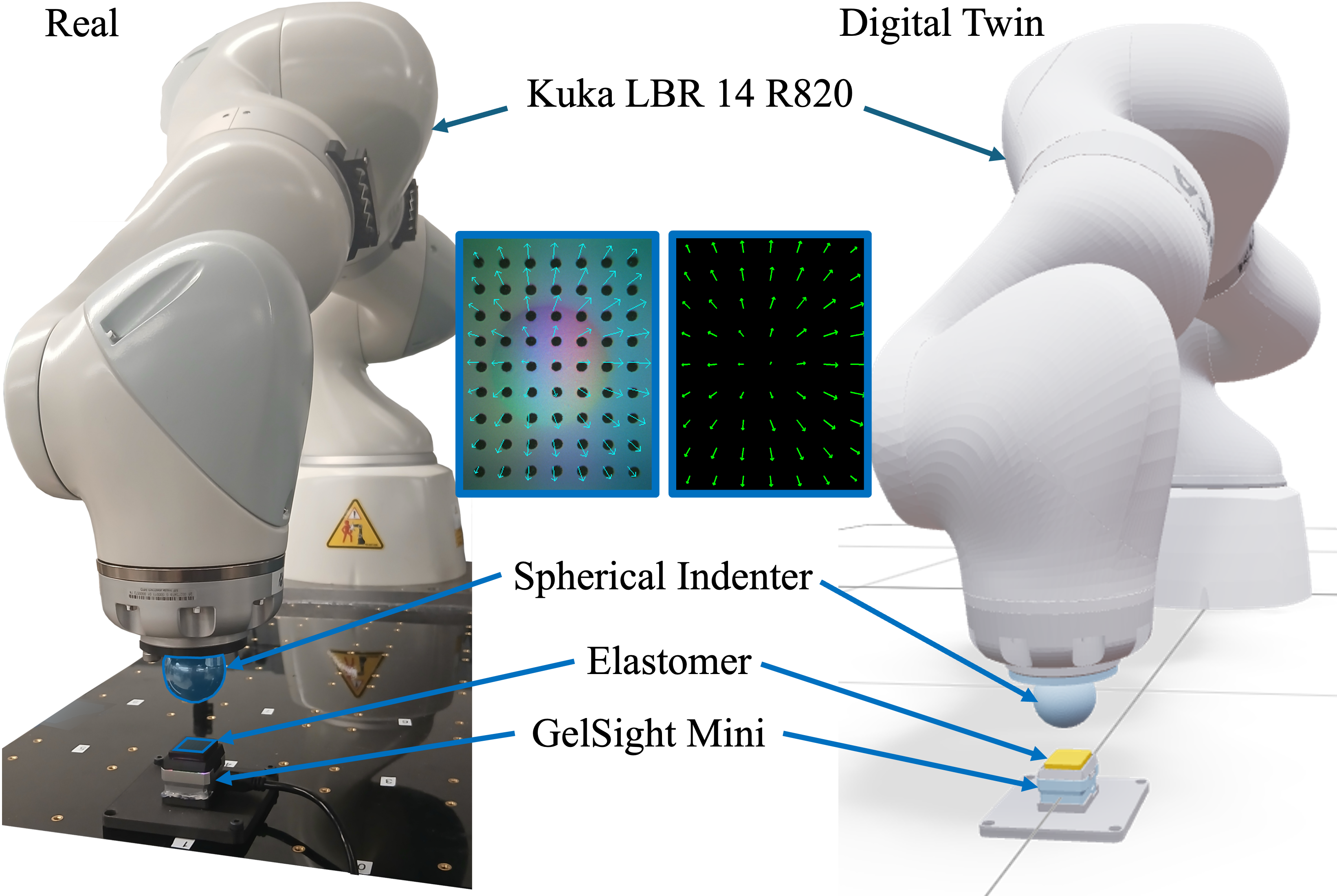}
    \caption{\textbf{Digital twin calibration setup:} In the real-world, a KUKA robot arm equipped with a sphere indenter to dilate and shear the elastomer of the GelSight Mini vision-based tactile sensor mounted on the table. We replicate the same motion in a digital twin of the real-world setup to run baseline and our algorithms to calibrate the tactile simulation model.}
    \label{fig:calibration}
    \vspace{-15pt}
\end{figure}

%% file: tables/marker_table.tex
\begin{table}[t]
\centering
\setlength{\tabcolsep}{3.5pt}
\begin{tabular}{lcccccccc}
\toprule
\multirow{4}{*}{Model} &
\multicolumn{8}{c}{Shear Type} \\
\cmidrule(lr){2-9}
& \multicolumn{2}{c}{Dilation}
& \multicolumn{2}{c}{Shear}
& \multicolumn{2}{c}{Twist}
& \multicolumn{2}{c}{Roll} \\
\cmidrule(lr){2-3}\cmidrule(lr){4-5}\cmidrule(lr){6-7}\cmidrule(lr){8-9}
& RMSE $\downarrow$ & CS $\uparrow$ & RMSE & CS & RMSE & CS & RMSE & CS \\
\midrule
TacSL
& 2.187 & 0.323
& 4.262 & 0.126
& 4.022 & 0.157
& 3.861 & 0.137 \\

FOTS
& 1.333 & \textbf{0.976}
& 2.146 & 0.930
& 3.596 & 0.683
& 2.841 & 0.561 \\

Ours
& \textbf{1.000} & \textbf{0.976}
& \textbf{1.719} & \textbf{0.951}
& \textbf{2.021} & \textbf{0.974}
& \textbf{1.576} & \textbf{0.904} \\
\bottomrule
\end{tabular}
\caption{\textbf{Calibration Results.} Per-taxel shear RMSE [px] and cosine similarity (denoted as CS) between the predicted and ground-truth shear vectors are reported for each shear type.}
\label{tab:calibration_results}
\vspace{-5mm}
\end{table}

%% file: floating/vedo_calibration.tex
\begin{figure}
    \centering
    \includegraphics[width=\linewidth]{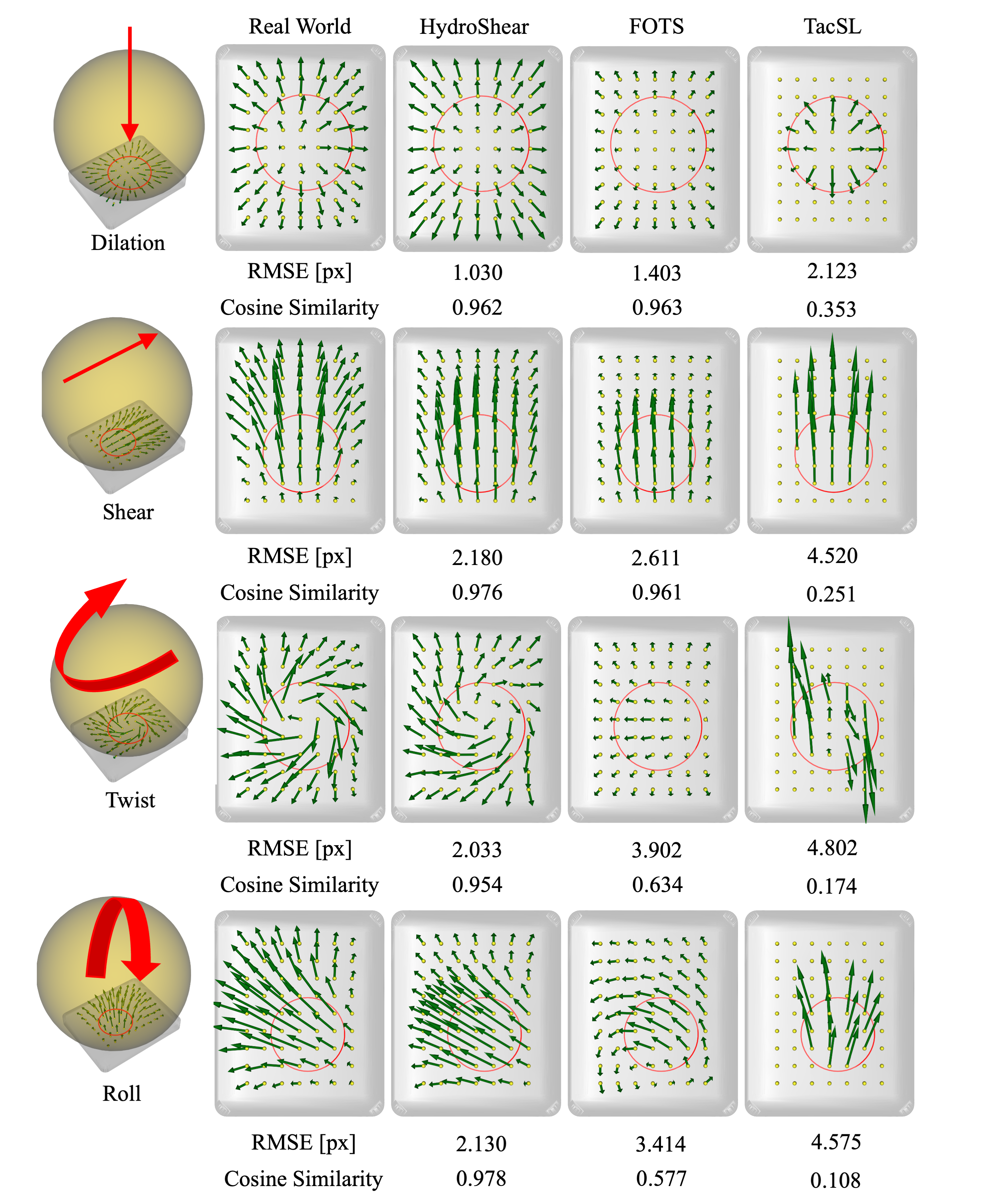}
    \caption{\textbf{Samples from calibration.} We showcase samples from representative motions such as dilation, translational shear, rotational shear (twist) and from rolling motion of the sphere. The red circle indicates the contact patch on the soft elastomer from the rigid sphere. The green arrows represent the shear vector field that are not to real-world scale but consistent across samples for visualization purposes. We list the RMSE and cosine similarity between the real-world ground truth sample and the simulated ones respectively as reference values to Tab.~\ref{tab:calibration_results}.}
    \label{fig:vedo_calibration}
\end{figure}

%% file: floating/vedo_complex_geometries.tex
\begin{figure}
    \centering
    \includegraphics[width=\linewidth]{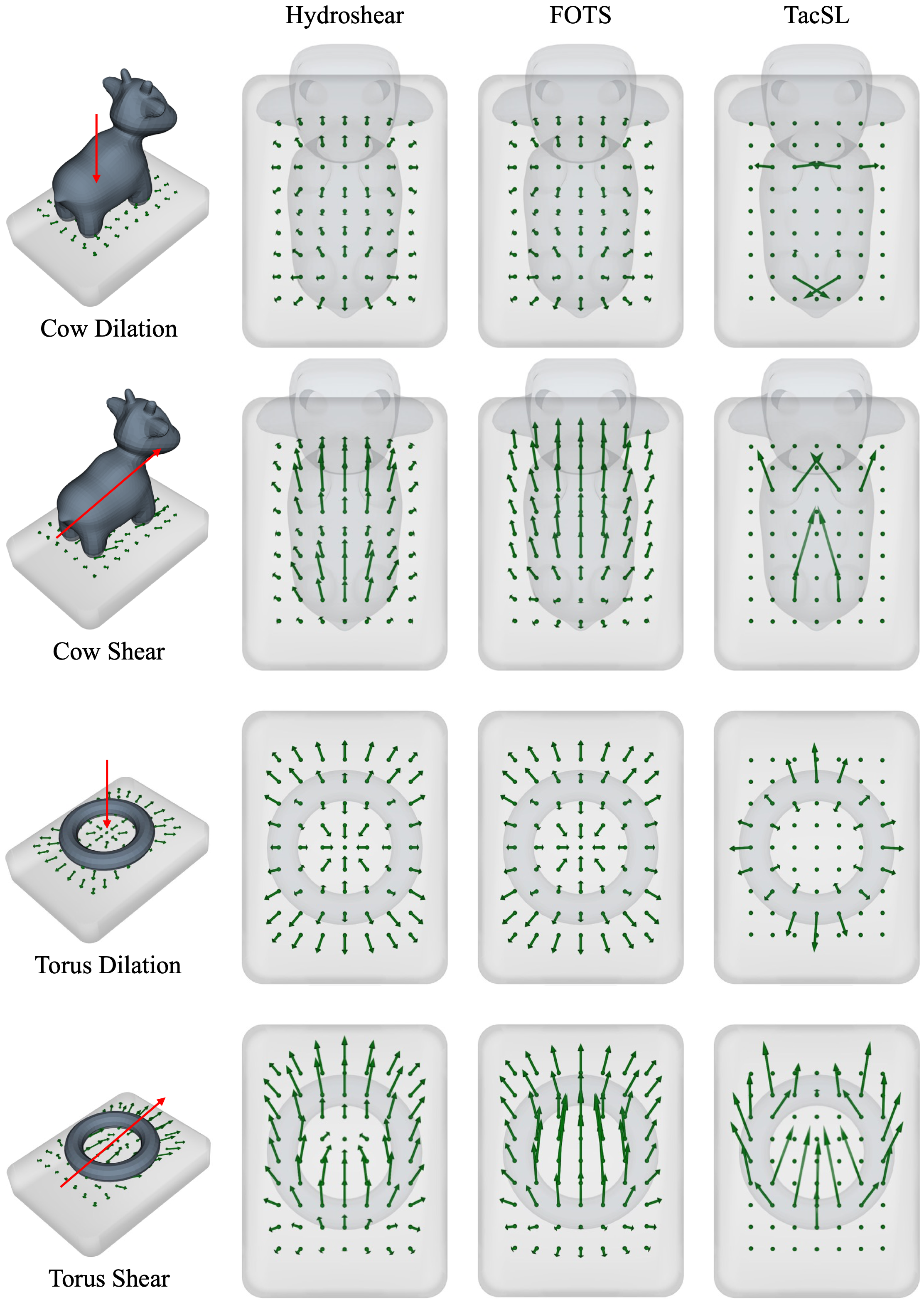}
    \caption{\textbf{Illustration of shear simulation with complex geometries.} We showcase \papername{}'s ability to simulate marker displacement fields from interactions between the sensor elastomer and complex geometries such as a cow with four legs and a ring torus. These geometries introduce complex configurations by the shape of the contact patch or by the number of it. We compare against the baseline models and find that with \papername{}'s SDF-based formulation and with tactile shadowing effects, we can effectively simulate the shear feedback resulting from complex contact geometries compared to other methods.}
    \label{fig:complex_geom}
\end{figure}

%% file: tables/zeroshot_table.tex
\begin{table}[h!]
\centering
\resizebox{\columnwidth}{!}{%
\begin{tabular}{lccccc}
\toprule
\textbf{Model} &
\begin{tabular}[c]{@{}c@{}} Peg \\ Insertion \end{tabular} &
\begin{tabular}[c]{@{}c@{}} Bin \\ Packing \end{tabular} &
\begin{tabular}[c]{@{}c@{}} Book \\ Shelving \end{tabular} &
\begin{tabular}[c]{@{}c@{}} Drawer \\ Pulling \end{tabular} &
\begin{tabular}[c]{@{}c@{}} Total \end{tabular} \\
\midrule
TacSL Gray & 16/30 & 16/30 & 6/30 & 3/30 & 41/120 \\ %
TacSL Shear & 19/30 & 4/30 & 23/30 & 24/30 & 70/120 \\ %
FOTS  & 1/30 & 5/30 & 20/30 & 15/30 & 41/120\\ %
FOTS (Reimpl.)  & 20/30 & 24/30 & 26/30 & 3/30 & 73/120\\ %
HydroShear  & \textbf{25/30} & \textbf{29/30} & \textbf{28/30} & \textbf{30/30} & \textbf{112/120} \\ %
\bottomrule
\end{tabular}%
}
\caption{\textbf{Zero-shot Sim-to-Real Policy Evaluation.} We test 5 different policies that is trained with different tactile simulation frameworks for each task: (1) TacSL tactile grayscale image;  (2) TacSL tactile normalized shear; (3) Original FOTS tactile shear with \textit{our} parallelization; (4) Our \textit{reimplemention} of FOTS with improvements; (5) HydroShear tactile shear.}
\label{tab:sim_to_real_results}
\end{table}

%% file: sections2/5_discussion_and_limitations.tex
\section{Discussion and Limitations}

\papername{} enables a powerful recipe for sim-to-real tactile policy transfer. While effective, our method depends on whether the underlying physics engine allows for penetrating contact simulation. In this work, we use Isaac Gym TacSL's Kelvin-Voigt implementation for compliant contact. Future extensions could incorporate physics engines that natively support hydroelastic contacts.

\papername{} uses SDFs to represent both the indenter and elastomer. Thus, using higher SDF resolutions can result in higher shear simulation accuracy at the cost of slowing down the simulation due to the increased number of points to track during indenter-elastomer penetration. In addition, SDFs are particularly effective for rigid-bodies. However, deformable objects may require an alternative representation that will affect the algorithms used to track points and compute forces. 

\papername{} assumes that the elastomer of the vision-based tactile sensor is flat. Future work could extend \papername{} to curved elastomers on dexterous hand fingertips.

\papername{} is effective for vision-based tactile sensors that rely on deformation cues. While extension to other tactile sensors is possible, their evaluation is outside the scope of the current work.

%% file: sections/acknowledgement.tex
We would like to thank Miquel Oller and James Lorenz for their help with the real-world experimental setup. This material is based upon work supported by the National Science Foundation under Grant No. 2337870 \& 2220876.

%% file: sections2/6_appendix.tex
\appendix

\subsection{FOTS}
\label{appendix:fots}

FOTS models the marker displacement field as the sum of three components:
\begin{align*}
\mathbf{M}_t( (x,y), {}^E\mathbf{X}^I_t ) = \mathbf{M}_t^d((x,y)) &+ \mathbf{M}_t^s((x,y), {}^E\mathbf{X}^{I}_t, {}^E\mathbf{X}^{I}_k) \\
&+ \mathbf{M}^r_t((x,y), {}^E\mathbf{X}^I_t, {}^E\mathbf{X}^I_k)
\end{align*}
where $\mathbf{M}_t$ denotes the total marker displacement field, while $\mathbf{M}_t^d$, $\mathbf{M}_t^s$, and $\mathbf{M}_t^r$ represent the dilation, shear, and twist components, respectively. Here, $(x,y) \in \mathbb{R}^2$ denotes a tactile query point on the 2D sensor plane. The dilation displacement field $\mathbf{M}^d_t$ follows the same formulation as Eq. \ref{eqn:dilation-vf} in the main text, where contact points are identified via the indenter SDF $\phi_I(\mathbf{p}_i) < 0$ and their influence is dissipated by a Gaussian kernel with parameter $\lambda_d$. This dilation component is shared across HydroShear and FOTS.

The shear displacement field $\mathbf{M}_t^s$ in FOTS is computed from the relative displacement $\Delta \mathbf{s}_t$ between the indenter poses at time steps $t$ and $k$, expressed in the elastomer frame. This displacement is applied uniformly across tactile query points and dissipated using a gaussian kernel centered at the 2D projection of the indenter position, denoted by $\mathbf{G}_t$. The parameter $\Delta s^{\text{max}}$ bounds the maximum shear magnitude induced by the indenter motion:
\begin{align*}
    \mathbf{M}^s_t &= \min\{1,  \frac{\Delta s^{\text{max}}}{\|\Delta \mathbf{s}_t \|} \} \Delta \mathbf{s}_t \cdot \exp \left\{ -\lambda_s \left\| \mathbf{G}_ t - \begin{bmatrix} x \\ y \end{bmatrix} \right\| \right\}  \\
    \Delta \mathbf{s}_t &= \mathbf{G}_t - \mathbf{G}_k \\
    \mathbf{G}_t &= \begin{bmatrix} 1 & 0 & 0 \\ 0 & 1 & 0 \end{bmatrix} {}^E\mathbf{X}^I_t \mathbf{0}_{3 \times 1}
\end{align*}

The twist displacement field $\mathbf{M}^r_t$ in FOTS comes from the yaw rotation of the indenter relative to the elastomer frame at time $t$ and $k$. Similar to the shear displacement field formulation in FOTS, $\exp\{-\lambda_r \| \mathbf{G}_t - \begin{bmatrix} x \\ y \end{bmatrix}\|\}$ dissipates the twist shear at a tactile query grid point based on its distance from $\mathbf{G}_t$ and $\Delta \theta^{\text{max}}$ bounds the amount of twist shear can occur due to the relative indenter rotation.
\begin{align*}
    \mathbf{M}^r_t &=  \left(\begin{bmatrix} \cos{\Delta \theta} & -\sin{\Delta \theta} \\
    \sin{\Delta \theta} & \cos{\Delta \theta} \end{bmatrix} - \mathbf{I}_{2 \times 2} \right) (\mathbf{G}_t - \begin{bmatrix} x \\ y \end{bmatrix}) \\
    &\cdot \exp\{-\lambda_r \| \mathbf{G}_t - \begin{bmatrix} x \\ y \end{bmatrix}\|\}\\
    \Delta \theta &= \min \{ 1, \left| \frac{\Delta \theta^\text{max}}{\theta_t - \theta_k} \right| \}(\theta_t - \theta_k)
\end{align*}

A key structural limitation of this formulation is that both $\mathbf{M}^s_t$ and $\mathbf{M}^r_t$ are computed relative to $\mathbf{G}_t$. When the object frame is poorly aligned with the actual contact patch the resulting shear field no longer reflects the local contact dynamics at the sensor surface. This misalignment is the primary source of the sim-to-real gap observed in the original FOTS results for Peg Insertion and Bin Packing (Tab. \ref{tab:sim_to_real_results}), and motivates the reimplementation described in Sec. \ref{sec:4B-zero-shot-sim-to-real}.

\subsection{Details on Tactile Sensors}

In this work, we use the Gelsight Mini vision-based tactile sensors in our experiments. We use dotted gels for shear-based policies and a non-dotted gel for image-based policies. We obtain the marker displacements by tracking the optical flow in pixels.

\section{Details on \papername{} RL Training}

\subsection{Actor-Critic Network Architecture}
\label{appendix:network_arch}
For both the actor and critic, we employ separate network streams with an identical backbone architecture, consisting of a two-layer LSTM with 256 hidden units, followed by an MLP with hidden dimensions $[512, 128, 64]$ and ELU activations. Modality-specific encoders are applied only at the input stage and vary depending on whether the policy is a privileged teacher actor or a student actor operating with observations available in the real world. The dimensionalities of privileged state inputs are listed in Tab.~\ref{tab:mdp_table_peg_insertion}, \ref{tab:mdp_summary_binpacking}, \ref{tab:mdp_summary_book_shelving}, \ref{tab:mdp_summary_drawer_pulling}.

\input{tables/mdp_tables}

For student actors, TacSL grayscale tactile images originally at resolution $[3 \times 320 \times 240]$ are downsampled to $[3 \times 80 \times 60]$. For shear inputs, we match the real-world GelSight Mini sensor's dotted gel configuration by using a $[2 \times 7 \times 9]$ marker grid for each of the left and right tactile sensors. These tactile feedback encoders are implemented as 2D convolutional networks with a spatial soft-argmax readout to retain spatial structure in tactile observations. For shear-based tactile force fields, we use a three-layer convolutional encoder with filter sizes $[32, 64, 64]$ and kernel size $3$ throughout. All layers use stride $1$, with padding applied in the first two layers and no padding in the final layer. For vision-based TacSL grayscale tactile inputs, we employ a three-layer convolutional encoder with filter sizes $[32, 64, 64]$ and kernel sizes $[8, 4, 3]$, respectively. The first convolution uses stride $2$, while the remaining layers use stride $1$, and no padding is applied in any layer. All convolutional encoders use ReLU activations and are followed by a spatial soft-argmax operation to produce compact feature representations before being passed to the shared LSTM–MLP backbone.

\subsection{Sim-to-Real Transfer Pipeline for Tactile RL}
\label{appendix:training_details}
\begin{itemize}
    \item Train a teacher actor-critic without any contact penalty such that it finds the most efficient solution.
    \item Finetune the teacher actor-critic with contact penalties to make it less aggressive while preserving the similar behavior and solutions to diverse configs.
    \item Distill a student actor using the finetuned teacher's critic with contact penalty. 
    \item Contact penalty is important because it makes the policy less aggressive, which prevents the GelSight from tearing.
\end{itemize}

Fig.~\ref{fig:rl_training_curves_teacher} presents the two-stage RL teacher training curves for all four tasks where the first stage is trained without any contact penalty and in the second stage, we finetune with contact penalties. Next, we present the RL training curves from AACD for all tasks in \ref{fig:rl_training_curves_peg_insertion}, \ref{fig:rl_training_curves_bin_packing}, \ref{fig:rl_training_curves_book_shelving}, \ref{fig:rl_training_curves_drawer_pulling}.

In these student training curves, we see that \papername{}'s performance and the baseline performance are on par although they have a huge sim-to-real performance gap. This is largely because highly accurate tactile physics can make the simulated learning problem harder by introducing localized stick-slip, path dependence, and richer variability. Simpler models may therefore yield smoother signals that are easier to exploit in simulation, but at the cost of a larger sim-to-real gap. \papername{} is designed to improve transfer fidelity in the real-world rather than performing well in simulation-only.

\input{tables/ppo_hyperparameters}

\input{tables/timesteps_to_success}

\subsection{Reward Functions for RL Training}

For all contact-rich insertion tasks (peg insertion, bin packing, and book shelving), our reward function can be formulated as the following:

\begin{align*}
    R = \alpha_{\text{keypoint}} \cdot r_{\text{keypoint}} +  \alpha_{\text{centered}} \cdot r_{\text{contact}} + r_{\text{action\_grad}}
\end{align*}
where $r_\text{keypoint}$ is the object-to-goal keypoint matching reward, $r_{\text{contact}}$ is the contact penalty, and $r_\text{action\_grad}$ is the action gradient penalty. $\alpha_{\text{keypoint}}$ is a scalar that increases if the object is not centered on the goal keypoints, and $\alpha_{\text{centered}}$ is a scalar that decreases the contact penalty if the object is centered.

The keypoint matching reward formulation used in Tab. (\ref{tab:mdp_summary}-\ref{tab:mdp_summary_drawer_pulling}) can be decomposed into the MSE component and the exponential component. This is written as follows:
\begin{align*}
    r_{\text{keypoint}} &= \sum_{i=1}^4 -\lambda_{\text{kpt}} e_i + \frac{\lambda_{\text{exp}}}{n + \exp\{m e_i\} + \exp\{-m e_i\}  }\\
    e_i &= \| \hat{\mathbf{y}}_i - \mathbf{y}_i \|_2^2
\end{align*}
where $\lambda_{\text{kpt}}$ and $\lambda_\text{exp}$ are the penalty weights for the MSE component and exponential component, respectively. The exponential component provides a localized shaping reward around low keypoint error, encouraging precise alignment between the grasped object and the goal pose. The parameters m and n control the sharpness and peak magnitude of this shaping term, respectively. $\hat{\mathbf{y}}_i$ and $\mathbf{y}_i$ are the ith keypoint of the grasped object and the goal. We use four collinear keypoints to represent each of the object and the goal pose. %

The contact penalty reward penalizes the pairwise contact force \(\mathbf{f}_{\mathrm{obj},b}^{(t)}\) between the manipulated object \(\mathrm{obj}\) and environment body \(b\) at timestep \(t\) for every environment body in the set of task-specific relevant bodies \(\mathcal{B}_{\mathrm{env}}^{\mathrm{task}}\). We define this reward as follows:
\[
r_{\mathrm{contact}}^{(t)}
=
-\lambda_{\mathrm{contact}}
\sum\nolimits_{b \in \mathcal{B}_{\mathrm{env}}^{\mathrm{task}}}
\|\mathbf{f}_{\mathrm{obj},b}^{(t)}\|_2.
\]
where $\lambda_{\text{contact}}$ is the contact-based penalty weight.
This discourages aggressive contacts (e.g. betw. peg-socket) by finetuning teacher with contact penalty. %

For tasks like peg insertion, bin packing, and book shelving, $\alpha_{\text{keypoint}}$ and $\alpha_{\text{centered}}$ are scale factors used to encourage the policy to insert only when it is aligned with the goal. For example in peg insertion, the policy could approach the socket from the sides instead of from the top. $\alpha_{\text{keypoint}}$ is set to 10 when the peg's z position is at the socket level but its xy-coordinates are not centered at the socket (approaching from side of socket). Otherwise, $\alpha_{\text{keypoint}}$ is set to 1. For $\alpha_{\text{centered}}$, it is set to 0.001 when the peg is centered and 1.0 when the peg isn't. This prevents the contact penalty reward from penalizing contact when the policy is inserting the peg. This heuristic is repeated for both bin packing and book shelving.

The action gradient penalty penalizes large changes in the action between timesteps and is formulated as follows:
\begin{align*}
    r_{\text{action\_grad}} = -\lambda_\text{action}\|\mathbf{a}_t - \mathbf{a}_{t-1}\|_2^2
\end{align*}
where $\mathbf{a}_t$ is the action at time t and $\lambda_\text{action}$ is the penalty weight for the action gradient.

For the drawer pulling task, the reward function is written as follows:
\begin{align*}
    R = r_{\text{keypoint}} + r_{\text{contact}} + r_{\text{action\_grad}} + r_{\text{perturb}}
\end{align*}
where $\alpha_\text{centered}$ and $\alpha_\text{keypoint}$ are removed, and $r_\text{perturb}$ is added. $r_\text{perturb}$ is the perturb penalty.

$r_\text{perturb}$ is the perturb penalty term that penalizes when the robot grasps too hard before any force perturbation occurs during training. This prevents the robot from always grasping as hard as it can and only grasping harder as a reaction to the force perturbation that occurs during training. This penalty is written as follows:
\begin{align*}
    r_{\text{perturb}} = \begin{cases}
        -\lambda_{\text{perturb}} & \text{grasp\_width} < \beta \text{ \& perturb} > 0 \\
        0 & \text{else}
    \end{cases}
\end{align*}
where $\lambda_{\text{perturb}}$ is the penalty scale and perturb refers to the force perturbation magnitude currently being applied to the drawer. $\beta$ is the grasp width threshold (in meters) below which the gripper applies sufficient force to resist slippage under the applied perturbation.

\input{tables/lambda_table}

\subsection{Task Environment Randomization}
\label{appendix:randomization_modes}

For each task, we apply several key task randomization modes to increase the dificulty of the task and to improve the sim-to-real transfer of our policies. In Peg Insertion, we randomize the socket position on the table and the in-hand grasp pose of the robot end-effector on the plug. In Bin Packing, we randomize the goal pose and the arrangements of the neighboring cubes such that they occlude the goal. The neighboring cubes are pushed towards the goal pose as seen in Fig.~\ref{fig:task_randomization_modes}. In Book Shelving, we only have one goal pose but we randomize the arrangements of the neighboring books to occlude the goal in two different ways. One way is to tilt one book to diagonally occlude the goal while the other is to push the neighboring books closer to the goal. For Drawer Pulling, we randomize the force perturbation magnitude and when to apply the force perturbation.

\input{floating/task_randomization_modes}

\input{floating/rl_training_curves}

\section{RL Reward Functions}
\label{app:rewards}

\subsection{Details on the Real-world Setup}

The Peg Insertion task uses an IndustReal board mounted on a Vention table approximately 0.5\,m from the Franka Emika Panda arm. The setup consists of a cylindrical peg and two sockets. One socket serves as the insertion target, while the second socket is used to keep the peg upright and fixed prior to grasping, enabling the robot to acquire the peg with a consistent in-hand pose.

The Bin Packing task uses the same IndustReal board setup together with 16 wooden cubes and custom 3D-printed bin walls. The bin walls are arranged such that the cubes fit tightly within a $4 \times 4$ grid centered on the board. Fifteen cubes are pre-arranged inside the bin with orientations aligned to the bin walls, while the remaining cube is placed outside the bin at a fixed location to enable consistent grasp acquisition by the robot.

For the Book Shelving task, a single-row bookshelf is mounted on top of the Vention table and populated with books of varying sizes. The target insertion location is marked on the shelf. Neighboring books are positioned to partially occlude the goal region, while additional bookends stabilize the remaining books and maintain the shelf configuration throughout evaluation.

For the Drawer Pulling task, we replace the default Franka hand with a Weiss WSG-50 gripper due to its ability to modulate grasp force more accurately and consistently. A drawer with a horizontal wooden handle is mounted at the edge of the Vention table. Additional weights are placed inside the drawer to vary the pulling resistance. The robot begins with a loose initial grasp (7\,mm width) that slips under perturbation and must continuously modulate the gripper width down to a minimum of 4\,mm to maintain stable contact during pulling.

\subsection{Rollouts and Comparison of Tactile Feedback}
\label{appendix:rollouts}

This section analyzes representative rollout keyframes from the four manipulation tasks: Peg Insertion (Fig.~\ref{fig:peginsertion_rollout_keyframes}), Bin Packing (Fig.~\ref{fig:binpacking_rollout_keyframes}), Book Shelving (Fig.~\ref{fig:bookshelving_rollout_keyframes}), and Drawer Pulling (Fig.~\ref{fig:drawerpulling_rollout_keyframes}). Across these tasks, we observe clear differences in how each tactile simulation method models contact-rich interactions.

When examining Drawer Pulling (Fig. \ref{fig:drawerpulling_rollout_keyframes}), we can see that FOTS cannot capture object slippage as well as \papername{}. FOTS does not model slippage as accurately as \papername{}, causing the predicted shear magnitude in the keyframes to continue increasing even after the drawer handle begins slipping relative to the gripper. 

Across all tasks, TacSL exhibits a notable absence of tactile shadowing effects. TacSL relies on a penalty-based approximation that estimates shear from penetration depth and relative velocity. As a result, deformations induced by one contact region do not affect neighboring taxels, leading to less realistic shear during contact-rich interactions.

The limitations of image-only tactile representations are most evident in the Book Shelving task (Fig.~\ref{fig:bookshelving_rollout_keyframes}). Under TacSL Grayscale, the tactile images contain little information regarding the contact interactions, object geometry, or object pose. The book surface is substantially larger than the fingertip, causing the elastomer to become fully covered during contact and obscuring localized textures and edge features across the sensor surface. Without these cues, policy performance degrades.

Overall, the qualitative comparisons in Fig.~\ref{fig:peginsertion_rollout_keyframes}--\ref{fig:drawerpulling_rollout_keyframes} highlight several key limitations of coarse tactile representations for dexterous manipulation. The inability of FOTS to accurately model slippage, the absence of tactile shadowing in TacSL, and the featureless tactile observations produced by TacSL Grayscale under full contact all contribute to the sim-to-real gap. In contrast, HydroShear models accurate stick-slip interactions, tactile shadowing, and path-dependent deformation, enabling high-fidelity tactile shear and robust zero-shot sim-to-real transfer across contact-rich manipulation tasks.

\subsection{Speed Comparison between FOTS, FOTS (Reimpl.), and \papername{}}

We compare the computational performance of FOTS~\cite{fots}, our reimplemented FOTS, and \papername{} by measuring the average compute time required to generate tactile shear fields during exemplar Peg Insertion rollouts. The rollouts are generated using a teacher policy trained for the Peg Insertion task. To evaluate scalability, we vary the number of simulated environments and measure how the runtime changes with increasing batch size. Tab.~\ref{tab:speed-table} summarizes the resulting compute times for each method.

HydroShear requires evaluating shear contributions across all active contact points \(|K_t|\) for each tactile query point, leading to a nominal computational complexity of \(O(N \times |K_t|)\), where \(N\) is the number of tactile query points. This is a higher nominal computational complexity than FOTS which has a computational complexity of \(O(N)\). However, both $\mathbf{M}_t^s$ and $\mathbf{M}_t^d$ are implemented using batched GPU operations, enabling parallelization across contact points and substantially reducing the effective runtime in practice. The reimplemented FOTS also benefits significantly from batched operations, leading to substantially lower runtime compared to the original method.
\input{tables/speed_table}

Both the reimplemented FOTS and \papername{} achieve substantially better computational performance than the original FOTS implementation due to increased parallelization. Among the three methods, the reimplemented FOTS exhibits the most consistent scaling behavior, with mean compute time varying by only approximately 1\,ms as the number of environments increases. This stability arises from its reliance almost entirely on batched tensor operations, which introduce minimal overhead as batch size increases.

In contrast, the original FOTS implementation does not leverage batched computation and therefore scales poorly with increasing numbers of environments. While \papername{} remains significantly faster than the original FOTS implementation, its runtime still increases with larger batch sizes. We find that this overhead primarily originates from the batched SDF computations required by HydroShear. Although these operations are parallelized, their computational cost still grows with the number of environments. Future improvements in batched SDF algorithms could therefore further improve the computational efficiency and scalability of \papername{}.

\subsection{Gravity Effect}\label{app:gravity-effect}

In the real world, when the robot grasps the object, the elastomer deforms due to the object's gravity. This results in a shear that's strictly due to the object's gravity. In IsaacGym, the object does not translate in-grasp unless it goes into contact with other objects which means \papername{} cannot capture this gravity effect. To bridge this sim-to-real gap, we implement an augmentation technique to capture this gravity effect by shifting the object pose while in-grasp and running hydroshear as follows:
\begin{align*}
    \mathbf{g}_{j,t} = F(\tilde{\mathbf{f}}_{j,t}, {}^{G}\mathbf{X}^E {}^E\mathbf{X}^I_t, {}^E\mathbf{X}^I_t, \bar{\mathbf{o}}_j; E, K, A_j, \mu)
\end{align*}
where $\mathbf{g}_{j,t}$ is the tactile shear from applying the augmentation transform ${}^{G}\mathbf{X}^E$ onto the current indenter pose ${}^E\mathbf{X}^I_t$. Note that $\tilde{\mathbf{f}}_{j,t+1}$ will not use $\mathbf{g}_{j,t}$ for its recursive calculation. This allows us to change the tactile shear due to pose augmentations without affecting the recursive calculations of \papername{}. In our evaluations, we use this for FOTS, our reimplemented FOTS, and \papername{} to capture the gravity effect.

\subsection{Point Cloud Density}\label{app:pcd-density}

Fig.~\ref{fig:pcd_density} analyzes the effect of point cloud density on both the tactile fidelity and computational performance of HydroShear. We evaluate different surfel sampling resolutions on the real-world calibration dataset and compare the resulting simulated shear fields against the measured real-world shear observations.

To densely cover the indenter surface, we sample points on the indenter mesh using surfels with diameter \(d = 1.5\) mm, chosen relative to the approximate \(2\) mm spacing between tactile grid points on the sensor surface. Increasing the surfel diameter reduces the number of sampled surface points and improves computational efficiency, but also decreases tactile fidelity due to the coarser geometric representation of the contact surface. Conversely, smaller surfels provide denser contact coverage and more accurate shear simulation at the cost of increased computation time.

\input{floating/point_cloud_ablation_figure}

\subsection{Calibration Shear Data Normalization}
\label{app:calibration-normalization}

As described in Sec.~\ref{method:calib}, the calibration of \(\lambda_s\) is performed separately from the calibration of the tangential stiffness parameter \(K\). However, both parameters influence the magnitude of the resulting shear field. Specifically, \(K\) converts incremental displacements into force vectors in \(\tilde{\mathbf{f}}_{j,t}\), while \(\lambda_s\) controls the spatial dissipation of tactile shadowing induced by these forces.

To better isolate the effect of tactile shadowing during \(\lambda_s\) calibration, we normalize the real-world shear observations \(\mathbf{Y}_i^s\) to match the mean and variance of the simulated shear magnitudes produced by \(\mathbf{M}_t\). This normalization reduces the influence of absolute force magnitude during optimization and allows the calibration procedure to focus primarily on matching the tactile shadowing structure of the shear field.

We first compute the mean and standard deviation of the real-world and simulated shear magnitudes:
\vspace{-2mm}
\begin{align*}
    \mu_y &= \tfrac{1}{N} \textstyle\sum \| \mathbf{Y}_i^s \|, \quad
    \sigma_y = \sqrt{\tfrac{1}{N}\textstyle\sum (\| \mathbf{Y}_i^s \| - \mu_y)^2} \\
    \mu_m &= \tfrac{1}{N} \textstyle\sum \| \mathbf{M}_i \|, \quad
    \sigma_m = \sqrt{\tfrac{1}{N}\textstyle\sum (\| \mathbf{M}_i \| - \mu_m)^2}.
\end{align*}

We then rescale the measured shear vectors as follows:
\vspace{-2mm}
\begin{align*}
    \hat{\mathbf{Y}}_i^s =
    \left[
    \left(
    \tfrac{\|\mathbf{Y}_i^s\| - \mu_y}{\sigma_y}
    \right)
    \sigma_m + \mu_m
    \right]
    \tfrac{\mathbf{Y}_i^s}{\|\mathbf{Y}_i^s\|}.
\end{align*}

This normalization preserves the directional structure of the measured shear field while aligning its magnitude statistics with the simulated shear distribution used during \(\lambda_s\) calibration.

\subsection{Friction Coefficient \texorpdfstring{$\mu$}{mu} Estimation Details}

As described in Sec.~\ref{method:calib}, the friction coefficient \(\mu\) governs the transition between sticking and slipping in the HydroShear formulation. We estimate \(\mu\) from the saturation behavior of the measured shear field during controlled translation experiments.

During these experiments, the indenter is translated tangentially across the elastomer surface while maintaining contact. Initially, the measured shear displacement increases approximately linearly as tangential forces accumulate under sticking contact. As the applied tangential force approaches the friction limit, the shear response begins to saturate and eventually plateaus, indicating the onset of slip.

We collect shear trajectories that capture both the pre-slip and post-slip regimes and optimize \(\mu\) to match the observed saturation behavior of the real-world shear field. This procedure enables HydroShear to reproduce realistic stick-slip transitions without requiring a dedicated force-torque sensor for friction estimation.

\input{floating/peginsertion_rollout_keyframes}
\input{floating/binpacking_rollout_keyframes}
\input{floating/bookshelving_rollout_keyframes}
\input{floating/drawerpulling_rollout_keyframes}
\input{floating/rollout_keyframes}

%% file: tables/mdp_tables.tex
\begin{table*}[!p]
\centering
\caption{Summary of our MDP for Peg Insertion, in terms of state, action, and reward components. Each component is denoted by its name, shorthand symbol, dimensionality and a brief description. $\dagger$: only used in simulation.}
\label{tab:mdp_summary}
\renewcommand{\arraystretch}{1.2}
\begin{tabular}{l c c l}
\hline
\textbf{State Component} & \textbf{Symbol} & \textbf{Dimension} & \textbf{Description} \\
\hline
Plug state$\dagger$        & $\mathbf{x}_t^{\text{plug}}$ & $\mathbb{R}^{13}$      & Plug pose (quaternion and position) and velocity \\
Robot state$\dagger$                 & $\mathbf{q}_t$ & $\mathbb{R}^{16}$      & Joint positions (panda joints and gripper joints) and velocities. \\
End-effector pose  & $\mathbf{X}_t^{\text{ee}}$ & $\mathbb{R}^{7}$   & Pose of the robot's end-effector \\
Socket pose                   & $\mathbf{X}_t^{\text{socket}}$   & $\mathbb{R}^{7}$       & Pose of the socket \\
Plug-Socket Contact Force$\dagger$ & $\mathbf{F}_t^{\text{plug-socket}}$ & $\mathbb{R}^3$ & Force vector due to the contact interaction between the plug and the socket \\
Plug-Left Elastomer Contact Force$\dagger$ & $\mathbf{F}_t^{\text{plug-left}}$ & $\mathbb{R}^3$ & Contact interaction forces between plug and the elastomer on left panda finger\\
Plug-Right Elastomer Contact Force$\dagger$ & $\mathbf{F}_t^{\text{plug-right}}$ & $\mathbb{R}^3$ & Contact interaction forces between plug and the elastomer on right panda finger\\
Physics parameters$\dagger$ & $\nu$   & $\mathbb{R}^{6}$       & Mass, friction, restitution of object and friction of robot, elastomer, and environment \\
\hline
\hline
\textbf{Observation Component} & \textbf{Symbol} & \textbf{Dimension} & \textbf{Description} \\
\hline
End-effector pose socket frame  & ${}^{\text{socket}}\mathbf{X}_t^{\text{ee}}$ & $\mathbb{R}^{7}$   & Pose of the robot's end-effector in socket frame. \\
Tactile Shear Field & $\mathbf{I}_t^{\text{shear}}$ & $\mathbb{R}^{H \times W \times 3}$ & Shear occuring at tactile grid points. \\
\hline
\hline
\textbf{Action Component} & \textbf{Symbol} & \textbf{Dimension} & \textbf{Description} \\
\hline
Delta End-effector Pose Action & $\Delta \mathbf{X}^{\text{ee}}$ & $\mathbb{R}^{7}$ & Desired transform to apply onto the current end-effector pose \\
Stiffness gains            & $\mathbf{K}_p$     & $\mathbb{R}^{6}$ & Cartesian Impedance Stiffness gains \\
Damping gains               & $\mathbf{K}_d$    & $\mathbb{R}^{6}$ & Cartesian Impedance Damping gains \\
\hline
\hline
\textbf{Reward Component} & \textbf{Symbol} & \textbf{Dimension} & \textbf{Description} \\
\hline
Task success reward         & $r_{\text{success}}$ & $\mathbb{R}$ & Reward for task success \\
Keypoint alignment penalty  & $r_{\text{keypoint}}$ & $\mathbb{R}$ & Negative mean distance between plug keypoints and socket keypoints \\
Action Derivative penalty   & $r_{\text{ad}}$ & $\mathbb{R}$ & Negative Norm of finite difference between current action and previous action. \\
Table Force penalty & $r_{\text{table}}$ & $\mathbb{R}$ & Sum of force norm between forces between table and all other objects. \\
Contact Force penalty & $r_{\text{contact}}$ & $\mathbb{R}$ & Sum of contact force norm between peg and socket scaled by $\lambda_{\text{contact}}$. \\
\hline
\end{tabular}
\label{tab:mdp_table_peg_insertion}
\end{table*}

\begin{table*}[!p]
\centering
\caption{Summary of our MDP for Bin Packing, in terms of state, action, and reward components. Each component is denoted by its name, shorthand symbol, dimensionality and a brief description. $\dagger$: only used in simulation.}
\label{tab:mdp_summary_binpacking}
\renewcommand{\arraystretch}{1.2}
\begin{tabular}{l c c l}
\hline
\textbf{State Component} & \textbf{Symbol} & \textbf{Dimension} & \textbf{Description} \\
\hline

Grasped Cube state$\dagger$ 
& $\mathbf{x}_t^{\text{cube}}$ 
& $\mathbb{R}^{13}$ 
& Grasped cube pose and velocity \\

Preset Cubes state$\dagger$ 
& $\mathbf{x}_{t}^{\text{preset-cubes}}$ 
& $\mathbb{R}^{7}$ 
& Pose information for 15 cubes inside the bin. \\

Robot state$\dagger$ 
& $\mathbf{q}_t$ 
& $\mathbb{R}^{14}$ 
& Joint positions and velocities \\

End-effector pose
& $\mathbf{X}_t^{\text{ee}}$ 
& $\mathbb{R}^{7}$ 
& Pose of the robot's end-effector \\

Goal pose
& $\mathbf{X}^{\text{goal}}$ 
& $\mathbb{R}^{7}$ 
& Target goal pose of grasped cube in end-effector frame \\

\multirow{4}{*}{Grasped Cube Contact Forces$\dagger$}
& $\mathbf{F}_t^{\text{cube-left}}$ 
& $\mathbb{R}^{3}$ 
& Contact force between grasped cube and left elastomer \\

& $\mathbf{F}_t^{\text{cube-right}}$ 
& $\mathbb{R}^{3}$ 
& Contact force between grasped cube and right elastomer \\

& $\mathbf{F}_t^{\text{cube-bin}}$ 
& $\mathbb{R}^{3}$ 
& Contact force between grasped cube and bin \\

& $\mathbf{F}_t^{\text{cube-presetcubes}}$ 
& $\mathbb{R}^{42}$ 
& Contact force between grasped cube and 15 preset cubes \\

Physics parameters$\dagger$
& $\nu$ 
& $\mathbb{R}^{6}$ 
& Mass, friction, restitution of object and friction of robot, elastomer, and env \\

\hline
\hline
\textbf{Observation Component} & \textbf{Symbol} & \textbf{Dimension} & \textbf{Description} \\
\hline

End-effector pose goal frame
& ${}^{\text{goal}}\mathbf{X}_t^{\text{ee}}$
& $\mathbb{R}^{7}$
& End-effector pose expressed in the goal frame \\

Tactile Shear Field
& $\mathbf{I}_t^{\text{tactile}}$
& $\mathbb{R}^{H \times W \times 3}$
& Shear occurring on tactile grid points \\

\hline
\hline
\textbf{Action Component} & \textbf{Symbol} & \textbf{Dimension} & \textbf{Description} \\
\hline

Delta End-effector Pose Action
& $\Delta \mathbf{X}^{\text{ee}}$
& $\mathbb{R}^{7}$
& Desired transform applied to the current end-effector pose \\

Stiffness gains
& $\mathbf{K}_p$
& $\mathbb{R}^{6}$
& Cartesian impedance stiffness gains \\

Damping gains
& $\mathbf{K}_d$
& $\mathbb{R}^{6}$
& Cartesian impedance damping gains \\

\hline
\hline
\textbf{Reward Component} & \textbf{Symbol} & \textbf{Dimension} & \textbf{Description} \\
\hline

Task success reward
& $r_{\text{success}}$
& $\mathbb{R}$
& Reward for task success \\

Keypoint alignment penalty
& $r_{\text{keypoint}}$
& $\mathbb{R}$
& Negative mean distance between grasped cube keypoints and goal keypoints \\

Action derivative penalty
& $r_{\text{ad}}$
& $\mathbb{R}$
& Negative norm of finite difference between consecutive actions \\

Table Force penalty & $r_{\text{table}}$ & $\mathbb{R}$ & Sum of force norm between forces between table and all other objects. \\
Contact Force penalty & $r_{\text{contact}}$ & $\mathbb{R}$ & Sum of contact force norm between grasped cube and the bin or other packed cubes. \\

\hline
\end{tabular}
\end{table*}

\begin{table*}[!p]
\centering
\caption{Summary of our MDP for Book Shelving, in terms of state, action, and reward components. Each component is denoted by its name, shorthand symbol, dimensionality and a brief description. $\dagger$: only used in simulation.}
\label{tab:mdp_summary_book_shelving}
\renewcommand{\arraystretch}{1.2}
\begin{tabular}{l c c l}
\hline
\textbf{State Component} & \textbf{Symbol} & \textbf{Dimension} & \textbf{Description} \\
\hline

Grasped Book state$\dagger$
& $\mathbf{x}_t^{\text{cube}}$
& $\mathbb{R}^{13}$
& Grasped book pose and velocity \\

Preset Books state$\dagger$
& $\mathbf{x}_{t}^{\text{preset-cube}}$
& $\mathbb{R}^{7}$
& Preset books information for 14 books inside the bookshelf \\

Robot state$\dagger$
& $\mathbf{q}_t$
& $\mathbb{R}^{16}$
& Joint positions (panda and gripper joints) and velocities \\

End-effector pose
& $\mathbf{X}_t^{\text{ee}}$
& $\mathbb{R}^{7}$
& Pose of the robot's end-effector \\

Goal pose
& $\mathbf{X}^{\text{goal}}$
& $\mathbb{R}^{7}$
& Target goal pose of grasped book \\

\multirow{3}{*}{Grasped Book Contact Forces$\dagger$}
& $\mathbf{F}_t^{\text{book-elastomer}}$
& $\mathbb{R}^{6}$
& Contact force between grasped book and both elastomers \\

& $\mathbf{F}_t^{\text{book-shelf}}$
& $\mathbb{R}^{3}$
& Contact force between grasped book and shelf \\

& $\mathbf{F}_t^{\text{book-presetbooks}}$
& $\mathbb{R}^{42}$
& Contact force between grasped book and 14 preset books \\

Physics parameters$\dagger$
& $\nu$
& $\mathbb{R}^{6}$
& Mass, friction, restitution of object and friction of robot, elastomer, and environment \\

\hline
\hline
\textbf{Observation Component} & \textbf{Symbol} & \textbf{Dimension} & \textbf{Description} \\
\hline

End-effector pose goal frame
& ${}^{\text{goal}}\mathbf{X}_t^{\text{ee}}$
& $\mathbb{R}^{7}$
& End-effector pose expressed in the goal frame \\

Tactile Shear Field
& $\mathbf{I}_t^{\text{tactile}}$
& $\mathbb{R}^{H \times W \times 3}$
& Shear occurring on tactile grid points \\

\hline
\hline
\textbf{Action Component} & \textbf{Symbol} & \textbf{Dimension} & \textbf{Description} \\
\hline

Delta End-effector Pose Action
& $\Delta \mathbf{X}^{\text{ee}}$
& $\mathbb{R}^{7}$
& Desired transform applied to the current end-effector pose \\

Stiffness gains
& $\mathbf{K}_p$
& $\mathbb{R}^{6}$
& Cartesian impedance stiffness gains \\

Damping gains
& $\mathbf{K}_d$
& $\mathbb{R}^{6}$
& Cartesian impedance damping gains \\

\hline
\hline
\textbf{Reward Component} & \textbf{Symbol} & \textbf{Dimension} & \textbf{Description} \\
\hline

Task success reward
& $r_{\text{success}}$
& $\mathbb{R}$
& Reward for task success \\

Keypoint alignment penalty
& $r_{\text{keypoint}}$
& $\mathbb{R}$
& Negative mean distance between grasped book keypoints and goal keypoints \\

Action derivative penalty
& $r_{\text{ad}}$
& $\mathbb{R}$
& Negative norm of finite difference between consecutive actions \\

Table Force penalty & $r_{\text{table}}$ & $\mathbb{R}$ & Sum of force norm between forces between table and all other objects. \\
Contact Force penalty & $r_{\text{contact}}$ & $\mathbb{R}$ & Sum of contact force norm between grasped book and the shelf or other books. \\

\hline
\end{tabular}
\end{table*}

\begin{table*}[!p]
\centering
\caption{Summary of our MDP for Drawer Pulling, in terms of state, action, and reward components. Each component is denoted by its name, shorthand symbol, dimensionality and a brief description. $\dagger$: only used in simulation.}
\label{tab:mdp_summary_drawer_pulling}
\renewcommand{\arraystretch}{1.2}
\begin{tabular}{l c c l}
\hline
\textbf{State Component} & \textbf{Symbol} & \textbf{Dimension} & \textbf{Description} \\
\hline
Handle state$\dagger$        & $\mathbf{x}_t^{\text{handle}}$ & $\mathbb{R}^{13}$      & Drawer handle pose and velocity \\
Drawer Box Joint State$\dagger$      & $\mathbf{q}_t^{\text{drawer}}$ & $\mathbb{R}^2$ & Drawer Prismatic Joint (for sliding the drawer) position and velocity. \\
Robot state$\dagger$                 & $\mathbf{q}_t$ & $\mathbb{R}^{16}$      & Joint positions (panda joints and gripper joints) and velocities. \\
End-effector pose  & $\mathbf{X}_t^{\text{ee}}$ & $\mathbb{R}^{7}$   & Pose of the robot's end-effector \\
Handle-Elastomer Contact Force & $\mathbf{F}_t^{\text{handle-elastomer}}$ & $\mathbb{R}^6$ & Contact interaction force vector between drawer handle and both elastomers \\
Handle-Finger Contact Force & $\mathbf{F}_t^{\text{handle-finger}}$ & $\mathbb{R}^3$ & Contact interaction force vector between drawer handle and both fingers \\
Physics parameters$\dagger$ & $\nu$   & $\mathbb{R}^{6}$       & Mass, friction, restitution of object and friction of robot, elastomer, and env \\
\hline
\hline
\textbf{Observation Component} & \textbf{Symbol} & \textbf{Dimension} & \textbf{Description} \\
\hline
End-effector pose  & $\mathbf{X}_t^{\text{ee}}$ & $\mathbb{R}^{9}$   & Pose of the robot's end-effector \\
Tactile Shear Field & $\mathbf{I}_t^{\text{tactile}}$ & $\mathbb{R}^{H \times W \times 3}$ & Shear occurring on tactile grid points. \\
\hline
\hline
\textbf{Action Component} & \textbf{Symbol} & \textbf{Dimension} & \textbf{Description} \\
\hline
Delta End-effector Pose Action & $\Delta \mathbf{X}^{\text{ee}}$ & $\mathbb{R}^{7}$ & Desired transform to apply onto the current end-effector pose \\
Gripper Joint Position Action & $\hat{\mathbf{q}}_t^{\text{gripper}}$ & $\mathbb{R}$ & Desired gripper position on robot to control how much to grasp the drawer handle \\
Stiffness gains            & $k_p$     & $\mathbb{R}^{6}$ & Cartesian Impedance Stiffness gains \\
Damping gains               & $k_d$    & $\mathbb{R}^{6}$ & Cartesian Impedance Damping gains \\
\hline
\hline
\textbf{Reward Component} & \textbf{Symbol} & \textbf{Dimension} & \textbf{Description} \\
\hline
Task success reward         & $r_s$ & $\mathbb{R}$ & Reward for task success \\
Keypoint alignment penalty  & $r_k$ & $\mathbb{R}$ & Negative mean distance between drawer handle keypoints and goal keypoints \\
Action Derivative penalty   & $r_{ad}$ & $\mathbb{R}$ & Negative finite difference between current action and previous action. \\
Table Force penalty & $r_{\text{table}}$ & $\mathbb{R}$ & Sum of force norm between contact forces between table and all other objects. \\
Pre-ForcePerturb Gripper Penalty & $r_{\text{preperturb}}$ & $\mathbb{R}$ & Grasp Penalty whenever the robot grasps more before force perturbation happens. \\
Post-ForcePerturb Gripper Reward & $r_{\text{postperturb}}$ & $\mathbb{R}$ & Sum of normal forces between elastomers and drawer handle during perturbation.  \\
Robot-Drawer Contact Penalty & $r_{\text{drawer}}$ & $\mathbb{R}$ &  Penalty term when robot contacts any part of drawer that's not the handle. \\
\hline
\end{tabular}
\label{tab:mdp_table_drawer_pulling}
\end{table*}

%% file: tables/ppo_hyperparameters.tex
\begin{table}[htbp]
\centering
\caption{Teacher PPO Hyperparameters.}
\begin{tabular}{l c}
\hline
\textbf{Hyperparameter} & \textbf{Value} \\
\hline
Discount factor & 0.99 \\
GAE parameter & 0.95 \\
Grad norm & 1.0 \\
Entropy coeff. & 0.0 \\
PPO clip range & 0.2 \\
Bounds loss coeff. & 0.0 \\
Policy loss coeff. & 1.0 \\
Value loss coeff. & 2.0 \\
Base learning rate & 1e-4 \\
Adaptive LR KL target & 2e-3 \\
Max. learning rate & 1e-2 \\
Num. environments & 1024 \\
Episode length & 256 \\
Mini-epochs & 4 \\
Minibatch size & 1024 \\
\hline
\end{tabular}
\end{table}

\begin{table}[htbp]
\centering
\caption{Student PPO Hyperparameters.}
\begin{tabular}{l c}
\hline
\textbf{Hyperparameter} & \textbf{Value} \\
\hline
Discount factor & 0.99 \\
GAE parameter & 0.95 \\
Grad norm & 1.0 \\
Entropy coeff. & 0.0 \\
PPO clip range & 0.2 \\
Bounds loss coeff. & 0.0 \\
Policy loss coeff. & 1.0 \\
Value loss coeff. & 2.0 \\
Base learning rate & 1e-4 \\
Adaptive LR KL target & 8e-3 \\
Max. learning rate & 1e-4 \\
Num. environments & 256 \\
Episode length & 256 \\
Mini-epochs & 4 \\
Minibatch size & 1024 \\
\hline
\end{tabular}
\end{table}

%% file: tables/timesteps_to_success.tex
\begin{table}[h!]
\centering
\resizebox{\columnwidth}{!}{%
\begin{tabular}{lcccc}
\toprule
\textbf{Model} &
\begin{tabular}[c]{@{}c@{}} \texttt{Peg} \\ \texttt{Insertion} \end{tabular} &
\begin{tabular}[c]{@{}c@{}} \texttt{Bin} \\ \texttt{Packing} \end{tabular} &
\begin{tabular}[c]{@{}c@{}} \texttt{Book} \\ \texttt{Shelving} \end{tabular} &
\begin{tabular}[c]{@{}c@{}} \texttt{Drawer} \\ \texttt{Pulling} \end{tabular} \\
\midrule
TacSL Gray &
76.9 / 74.0 &
96.9 / 89.5 &
178.2 / 176.5 &
\textbf{63.7} / \textbf{63.0} \\

TacSL Shear &
87.1 / 81.0 &
97.7 / 91.0 &
\textbf{99.4} / \textbf{93.0} &
155.6 / 125.0 \\

FOTS &
197.0 / 197.0 &
237.6 / 213.0 &
158.2 / 163.0 &
96.8 / 90.0 \\

FOTS (Reimpl.) &
\textbf{74.9} / \textbf{73.5} &
169.2 / 148.5 &
126.7 / 114.0 &
75.7 / 72.0 \\

HydroShear &
85.2 / 82.0 &
108.8 / 86.0 &
180.2 / 182.5 &
120.6 / 116.5 \\
\bottomrule
\end{tabular}%
}
\caption{\textbf{Mean / median real-world environment timesteps to success (success-only). Lower is better.}}
\label{tab:real_world_timesteps}
\end{table}

%% file: tables/lambda_table.tex
\begin{table}[htbp]
\centering
\caption{Reward parameters used in training. $\lambda_\text{contact}$ is split into two parts where there is $\lambda_\text{contact}^\text{stage\_1}$ and $\lambda_\text{contact}^\text{stage\_2}$ for the stage 1 and stage 2 teacher training.}
\begin{tabular}{l c}
\hline
Parameters & \textbf{Value} \\
\hline
m & 50 \\
n & 0.0001 \\
$\beta$ & 0.0029 \\
$\lambda_\text{contact}^\text{stage\_1}$ & 0 \\
$\lambda_\text{contact}^\text{stage\_2}$ & 0.001 \\
$\lambda_\text{action}$ & 0.001 \\
$\lambda_\text{exp}$ & 0.1 \\
$\lambda_\text{kpt}$ & 0.1 \\
$\lambda_\text{perturb}$ & 0.1 \\
\hline
\end{tabular}
\end{table}

%% file: floating/task_randomization_modes.tex
\begin{figure}
    \centering
    \includegraphics[width=\linewidth]{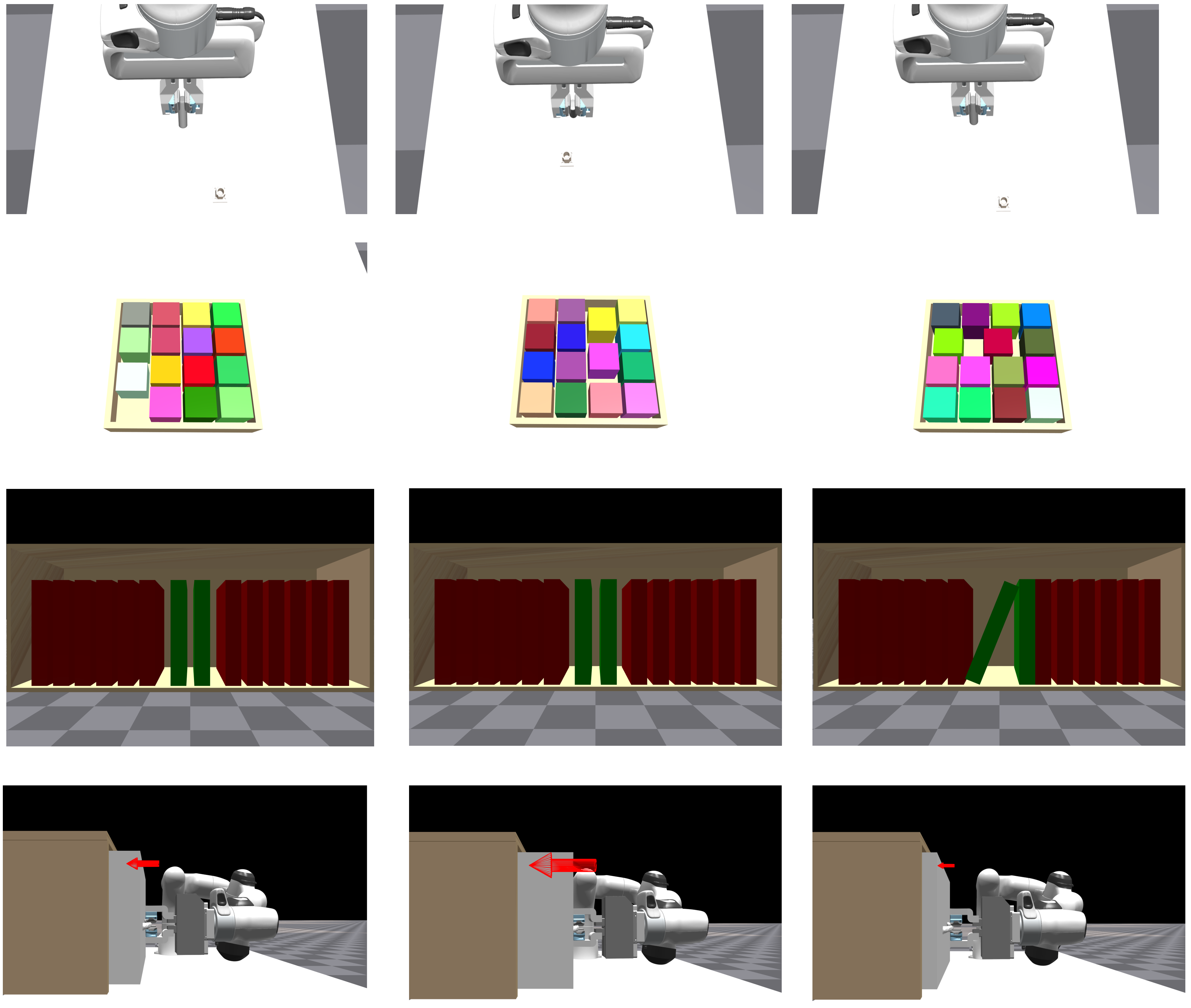}
    \caption{\textbf{Different randomization modes per task.} For Peg Insertion, we randomize the socket position and the in-hand pose of the peg. For Bin Packing, we randomize the goal location out of 16 cube locations and randomize the amount of squish, its direction (horizontal or vertical), and the number of cubes (single or double) used to squish the goal space. For Book Shelving, we randomize the books that are neighboring the goal book pose. We either squish the goal by translating neighboring books in parallel or by tilting to occlude the goal region. For Drawer Pulling, we randomize the force and timing at which perturbation is applied to induce slippage as the robot policy pulls the drawer out.}
    \label{fig:task_randomization_modes}
\end{figure}

%% file: floating/rl_training_curves.tex
\begin{figure}
    \centering
    \includegraphics[width=\linewidth]{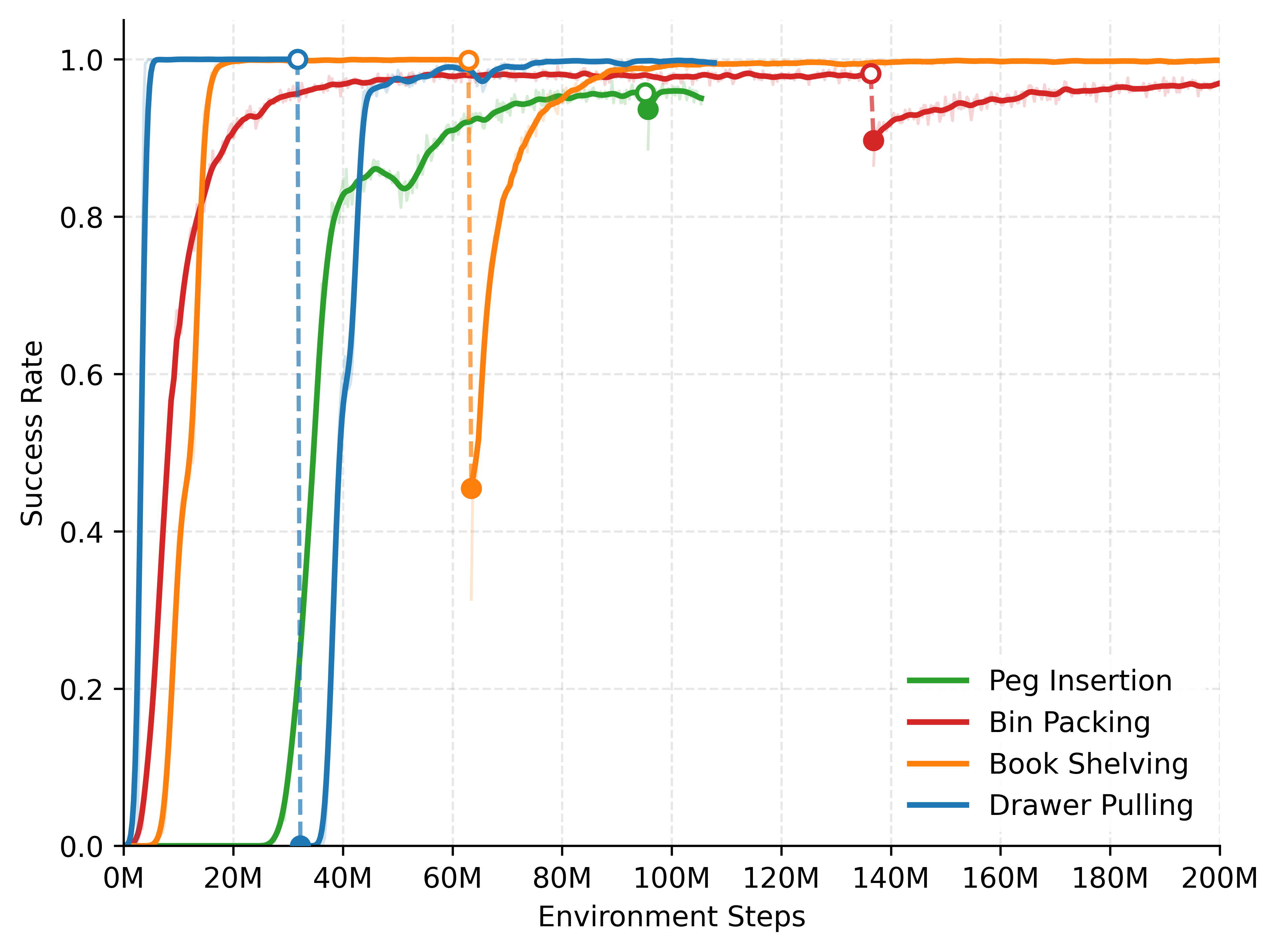}
    \caption{Teacher RL training curves. We train the teacher actor-critic over two stages: one without contact penalty and one with contact penalty. The discontinuity is from this finetuning.}
    \label{fig:rl_training_curves_teacher}
\end{figure}

\begin{figure}
    \centering
    \includegraphics[width=\linewidth]{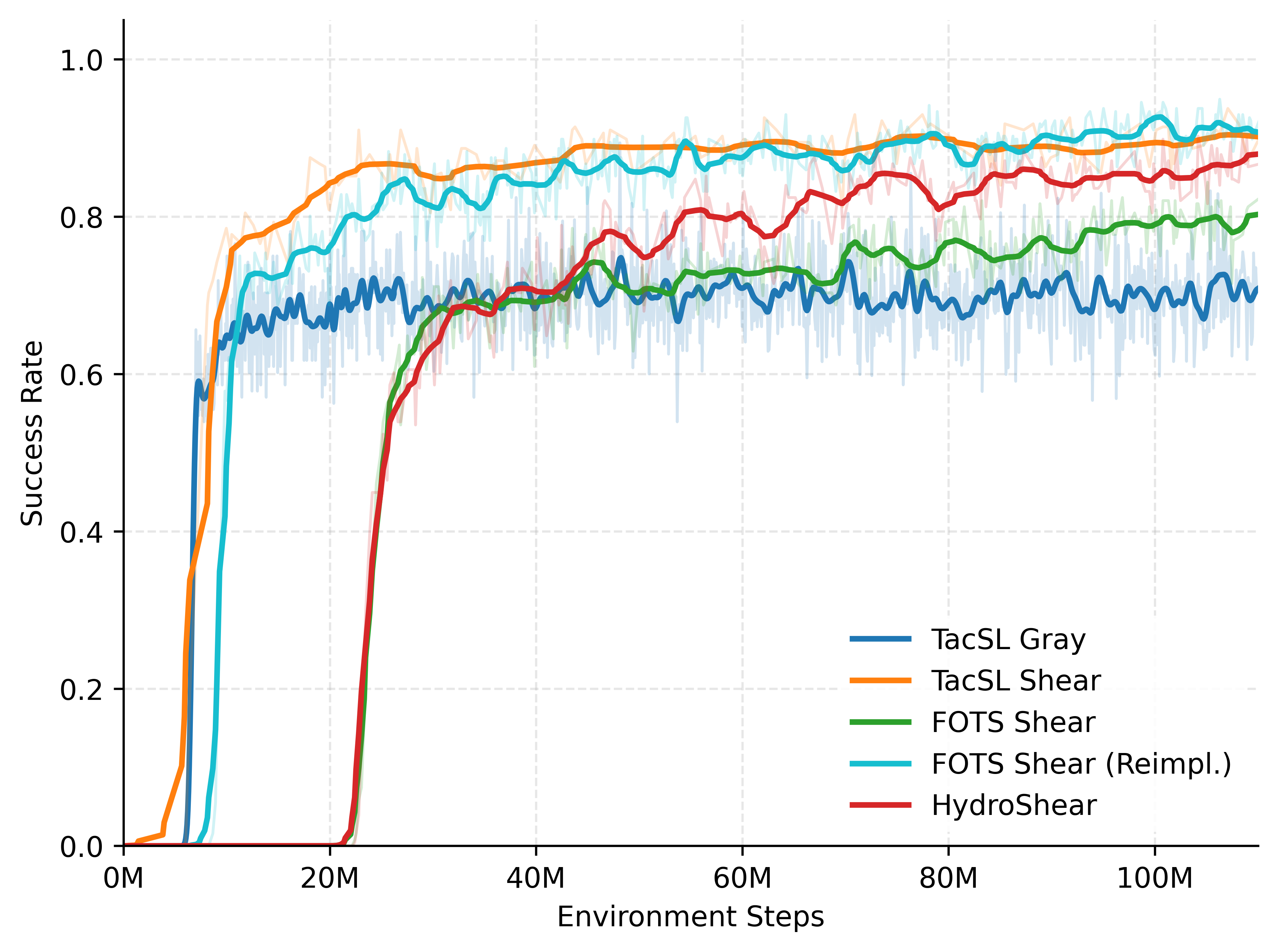}
    \caption{Peg Insertion Students RL training curves}
    \label{fig:rl_training_curves_peg_insertion}
\end{figure}

\begin{figure}
    \centering
    \includegraphics[width=\linewidth]{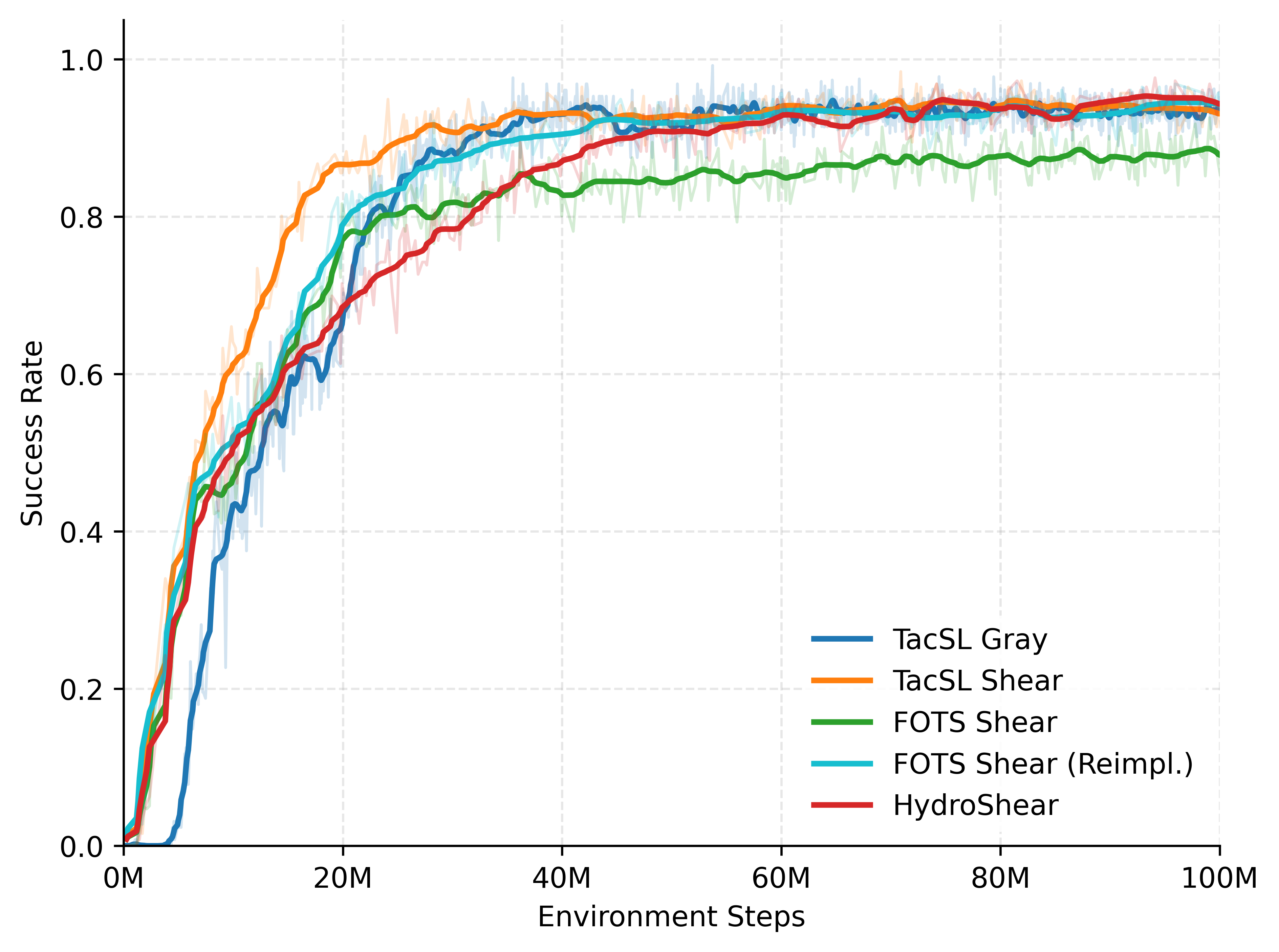}
    \caption{Bin Packing Students RL training curves}
    \label{fig:rl_training_curves_bin_packing}
\end{figure}

\begin{figure}
    \centering
    \includegraphics[width=\linewidth]{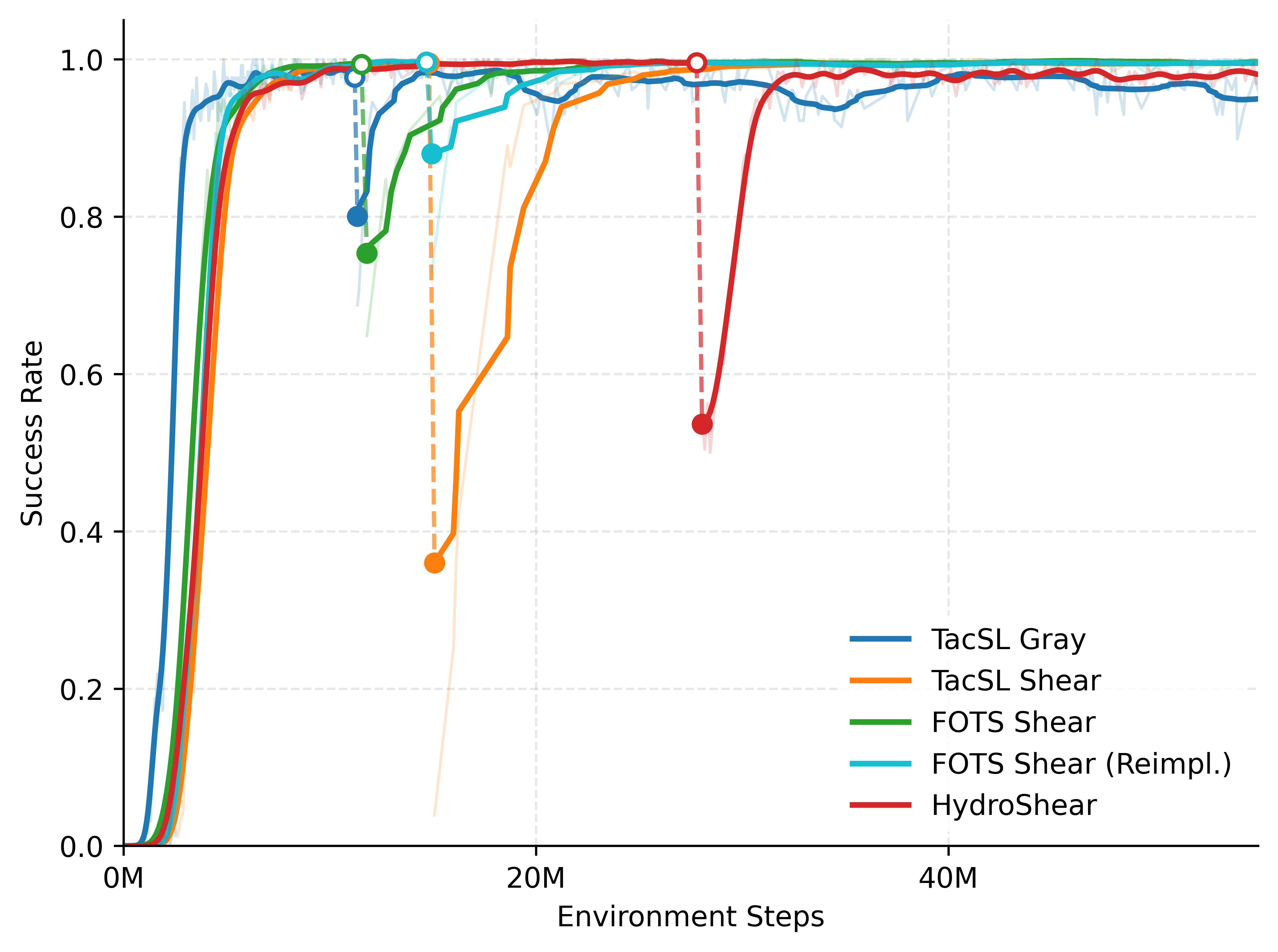}
    \caption{Book Shelving Students RL training curves}
    \label{fig:rl_training_curves_book_shelving}
\end{figure}

\begin{figure}
    \centering
    \includegraphics[width=\linewidth]{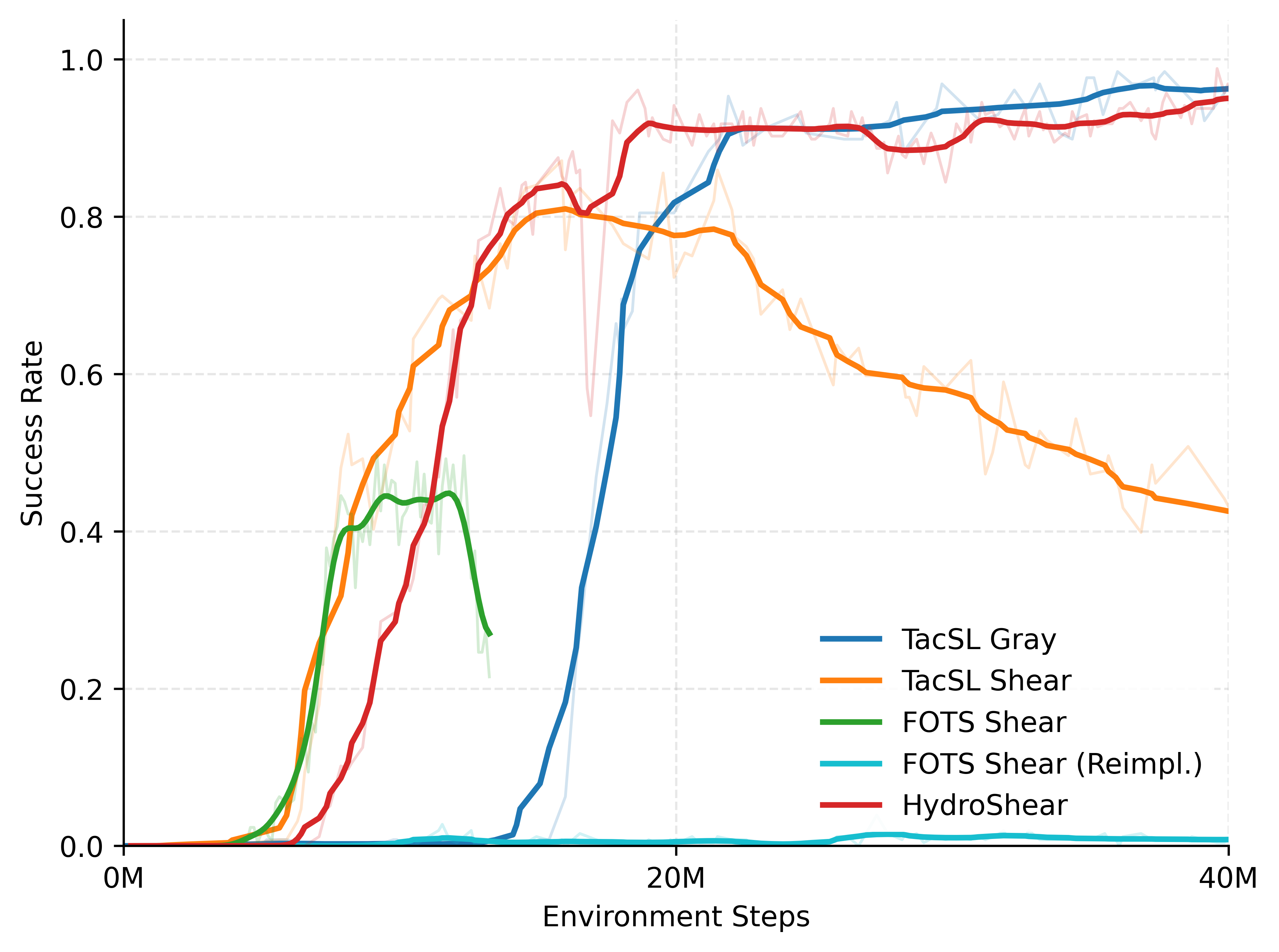}
    \caption{Drawer Pulling Students RL training curves}
    \label{fig:rl_training_curves_drawer_pulling}
\end{figure}

%% file: tables/speed_table.tex
\begin{table}[t]
\centering
\setlength{\tabcolsep}{3.5pt}
\begin{tabular}{lcccccccc}
\toprule
\multirow{2}{*}{Sensor Model} &
\multicolumn{6}{c}{Computation Time (ms)} \\
\cmidrule(lr){2-7}
& \multicolumn{2}{c}{256 Env}
& \multicolumn{2}{c}{512 Env}
& \multicolumn{2}{c}{1024 Env} \\
\cmidrule(lr){2-3}\cmidrule(lr){4-5}\cmidrule(lr){6-7}
& Mean $\downarrow$ & Stdev $\uparrow$ & Mean & Stdev & Mean & Stdev \\
\midrule
FOTS
& 122.829 & 3.445
& 230.884 & 6.360
& 446.656 & 4.028 \\

FOTS (Reimpl.)
& 11.145 & 1.826
& 11.827 & 1.621
& 12.040 & 1.593 \\

Ours
& 33.805  & 1.505
& 57.042  & 1.787
& 103.736 & 1.569\\
\bottomrule
\end{tabular}
\caption{\textbf{Comptutation Speed Comparison.} We evaluate the computational speed of FOTS, our implemented FOTS, and \papername{} by running it on the rollouts of a teacher policy on a Peg Insertion task. The number of environments labelled in the table describes the number of rollouts that the model has to calculate the tactile shear for during the evaluation. The computation time mean and standard deviation are calculated in milliseconds.}
\label{tab:speed-table}
\vspace{-5mm}
\end{table}

%% file: floating/point_cloud_ablation_figure.tex
\begin{figure}[H]
    \centering
    \includegraphics[width=1.0\linewidth]{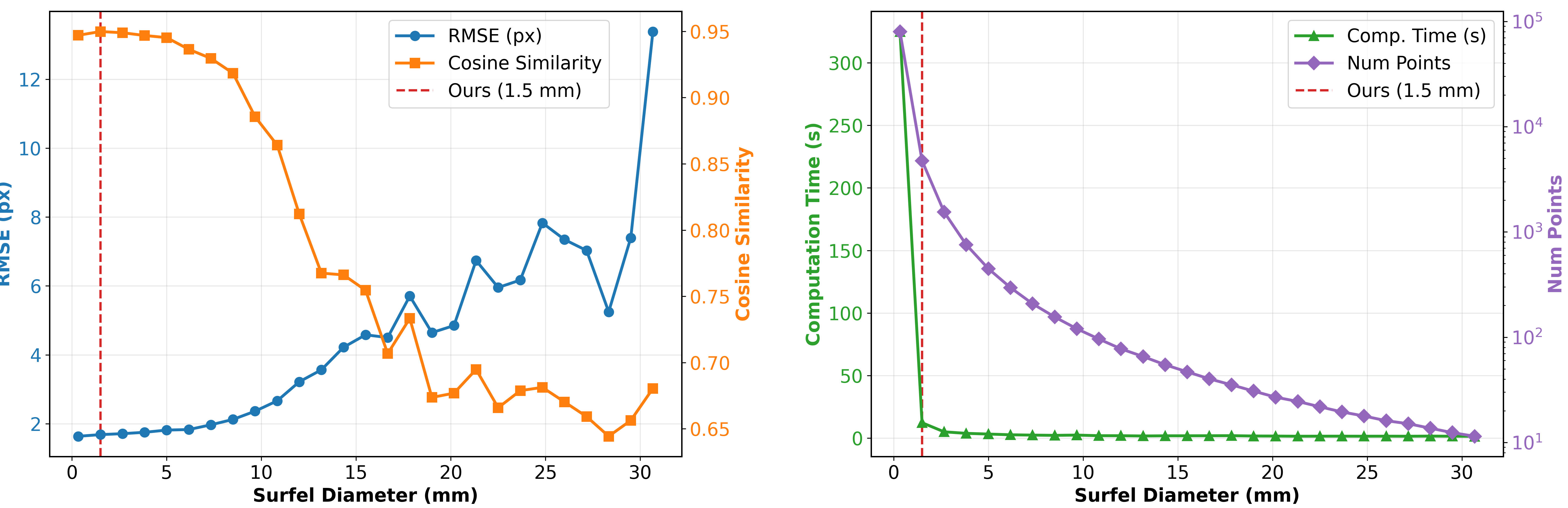}
    \caption{The effect of higher mesh point density (smaller surfel diameter) on HydroShear performance. We use 1.5mm surfels.}
    \label{fig:pcd_density}
\end{figure}

%% file: floating/peginsertion_rollout_keyframes.tex
\begin{figure*}[!p]
    \centering
    \includegraphics[width=1.00\linewidth]{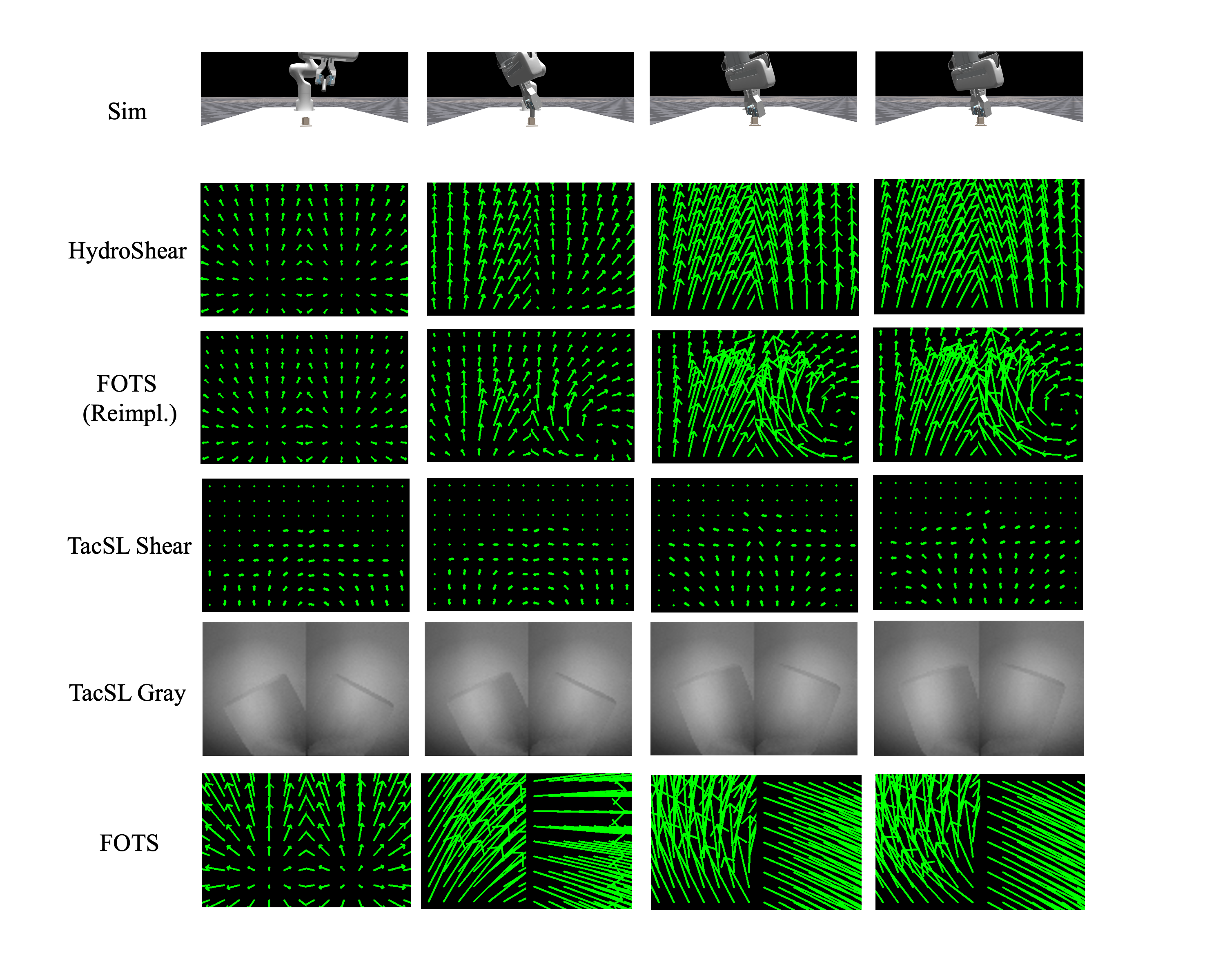}
    \caption{Illustration of representative keyframes during the same policy rollout for peg insertion and its respective tactile feedback per simulation framework.}
    \label{fig:peginsertion_rollout_keyframes}
\end{figure*}

%% file: floating/binpacking_rollout_keyframes.tex
\begin{figure*}[!p]
    \centering
    \includegraphics[width=1.00\linewidth]{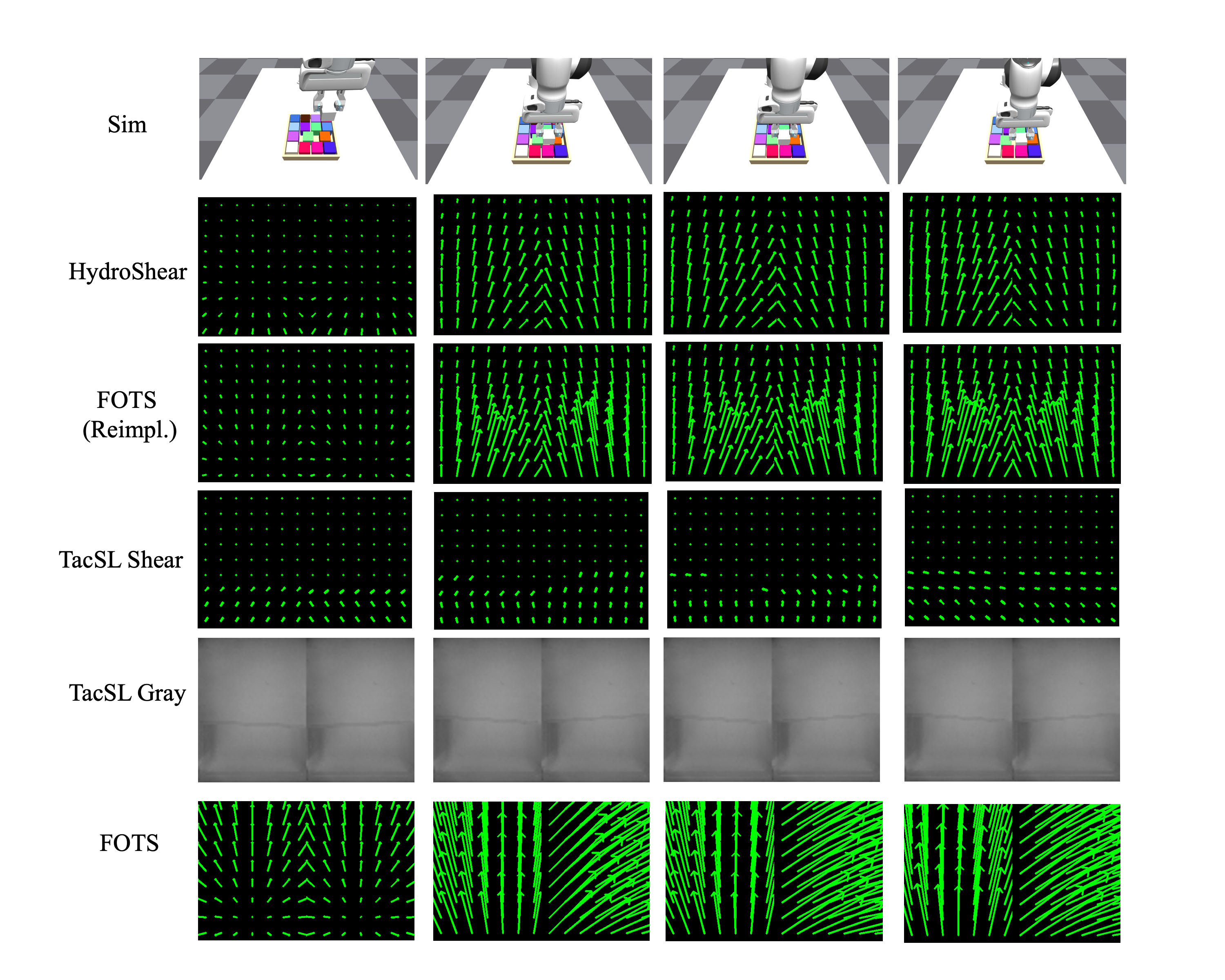}
    \caption{Illustration of representative keyframes during the same policy rollout for bin packing and its respective tactile feedback per simulation framework.}
    \label{fig:binpacking_rollout_keyframes}
\end{figure*}

%% file: floating/bookshelving_rollout_keyframes.tex
\begin{figure*}[!ht]
    \centering
    \includegraphics[width=0.90\linewidth]{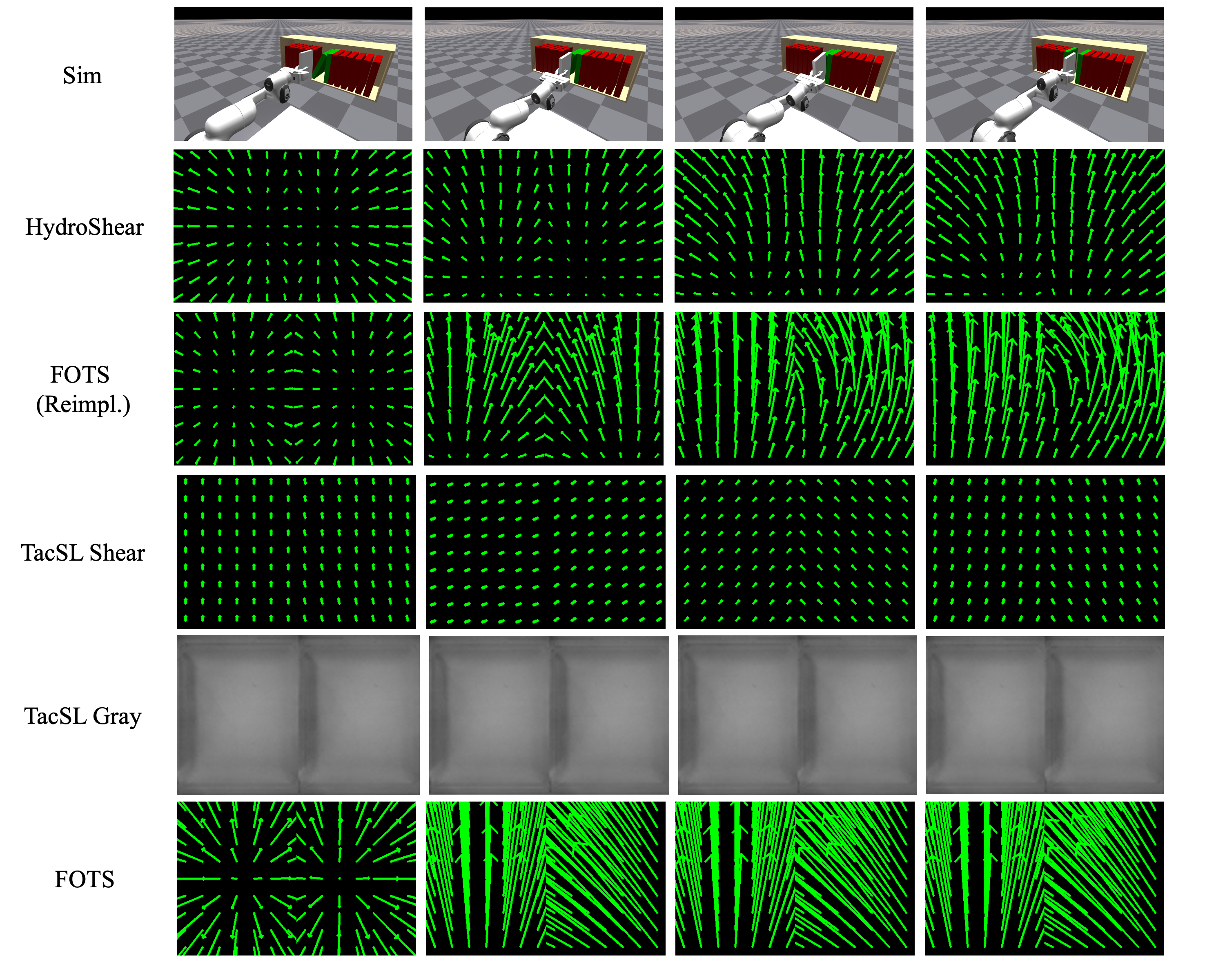}
    \caption{Illustration of representative keyframes during the same policy rollout for book shelving and its respective tactile feedback per simulation framework.}
    \label{fig:bookshelving_rollout_keyframes}
\end{figure*}

%% file: floating/drawerpulling_rollout_keyframes.tex
\begin{figure*}[!p]
    \centering
    \includegraphics[width=1.00\linewidth]{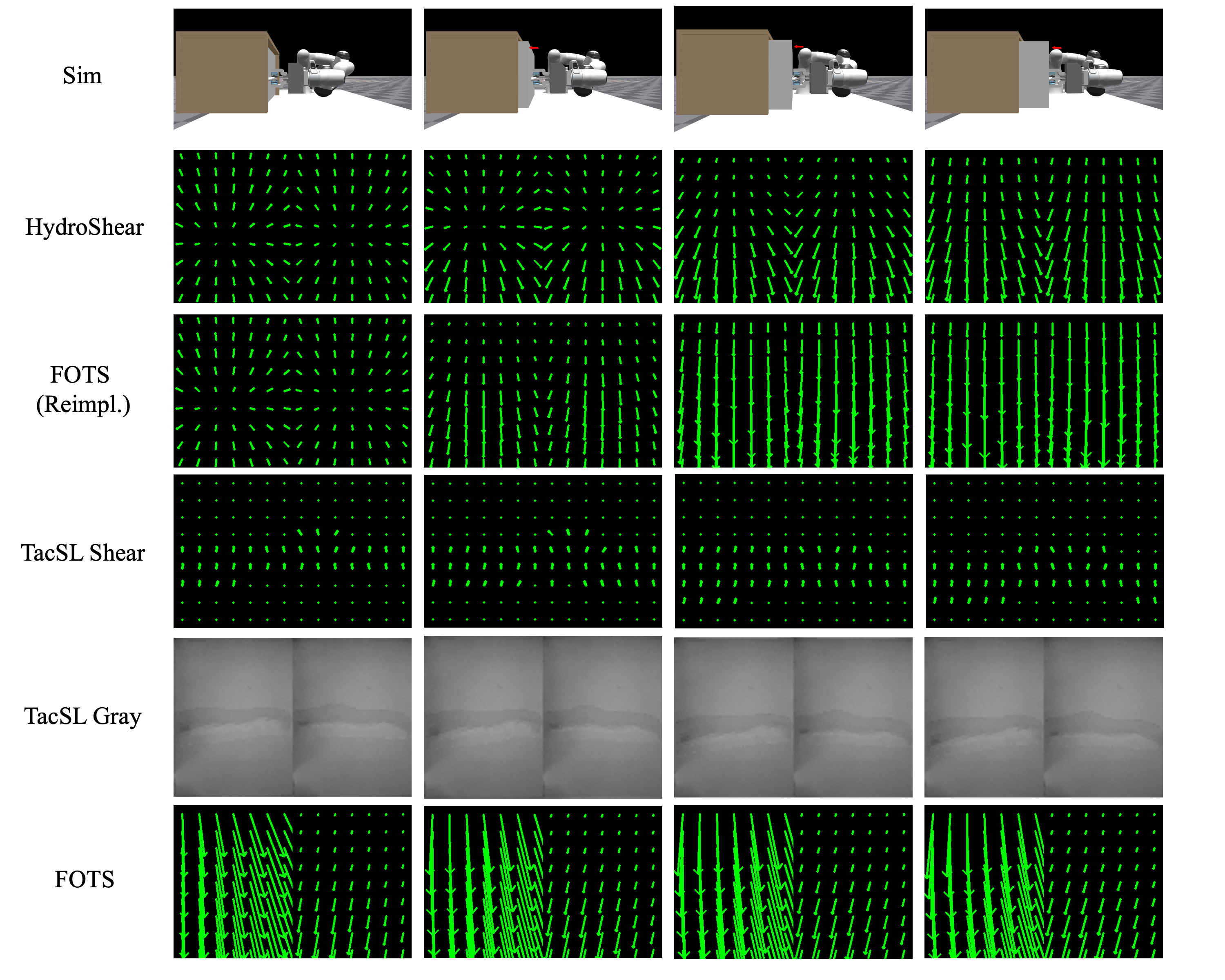}
    \caption{Illustration of representative keyframes during the same policy rollout for drawer pulling and its respective tactile feedback per simulation framework.}
    \label{fig:drawerpulling_rollout_keyframes}
\end{figure*}

%% file: floating/rollout_keyframes.tex
\begin{figure*}[!p]
    \centering
    \includegraphics[width=0.90\linewidth]{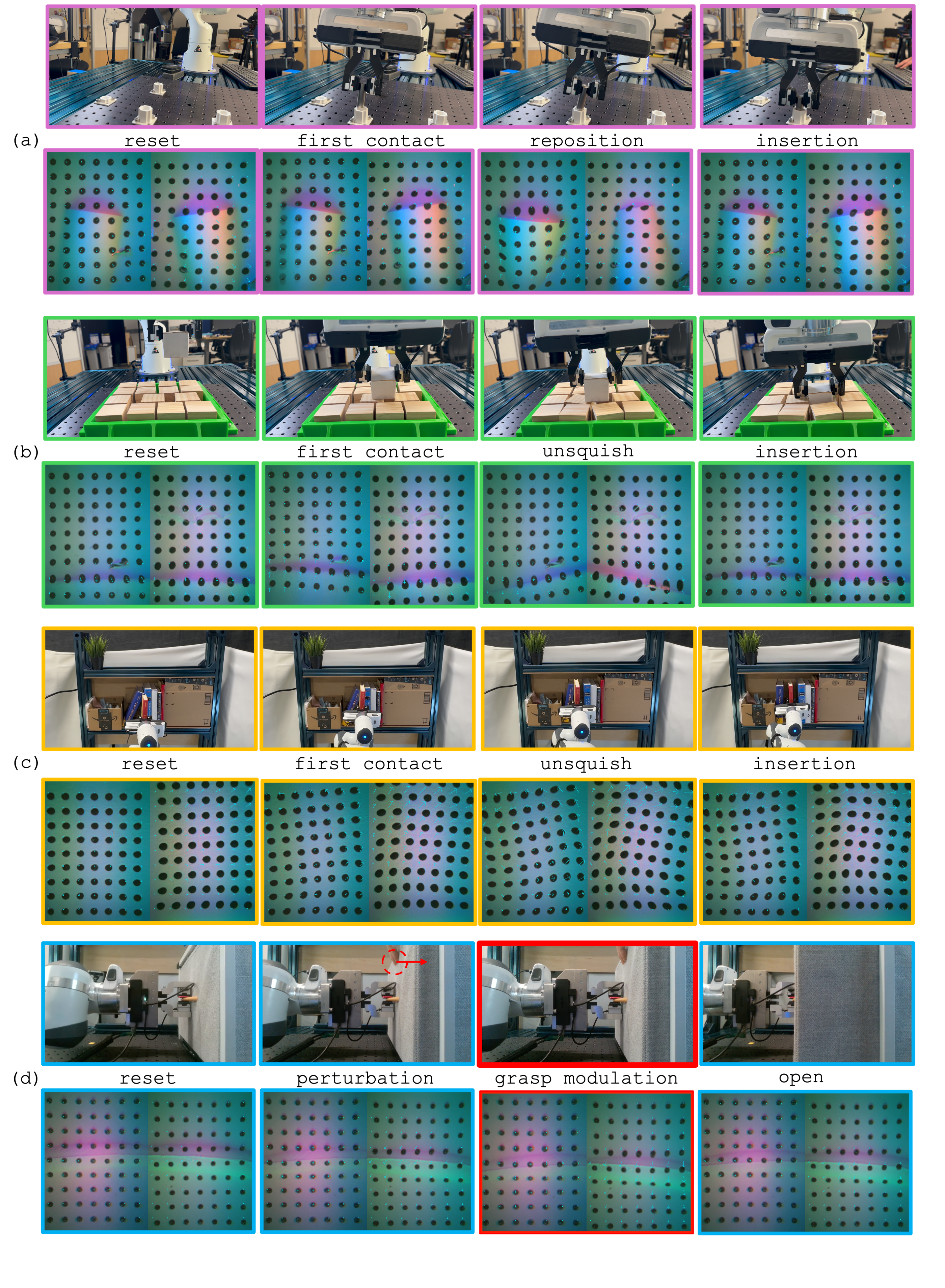}
    \caption{Illustration of representative keyframes from a sim-to-real policy rollout for each task: (a) Peg Insertion, (b) Bin Packing, (c) Book Shelving, and (d) Drawer Pulling. For (a–c), we show the reset frame, the moment when the grasped object first makes extrinsic contact with the environment, the policy’s reactive response to this contact, and the final successful insertion. For (d), we show the reset with the initial grasp, a frame where a random force perturbation is applied, a frame where the gripper adaptively modulates its grasp to prevent further slippage (highlighted by the red outline), and the final successful drawer opening while maintaining a stable grasp. For each task, the top row shows the visual scene during the rollout, while the bottom row shows synchronized tactile images overlaid with marker displacement fields, visualized as light blue arrows.}
    \label{fig:rollout_keyframes}
\end{figure*}